\documentclass[final,1p,times]{elsarticle}

\usepackage{graphicx}
\usepackage{amssymb}
\usepackage{amsthm}
\usepackage{amsmath}
\usepackage{lineno}
\usepackage{pifont}
\usepackage{multirow}
\usepackage{multicol}
\usepackage{subcaption}
\usepackage{comment}
\usepackage{booktabs}

\journal{Elsevier}

\newenvironment{myalign}%
  {\linenomath\align}%
  {\endalign\endlinenomath}

\begin{document}

\begin{frontmatter}


\title{Dense open-set recognition\\
 based on training with noisy negative images}


\author{Petra Bevandić,
        Ivan Krešo,
        Marin Oršić,
        and Siniša Šegvić}

\address{Faculty of Electrical Engineering and Computing,
University of Zagreb, Croatia}

\begin{abstract}
Deep convolutional models often produce 
inadequate predictions
for inputs which are foreign 
to the training distribution.
Consequently, the problem of 
detecting outlier images
has recently been receiving a lot of attention.
Unlike most previous work,
we address this problem in 
the dense prediction context.
Our approach is based on two reasonable assumptions.
First, we assume that the inlier dataset is related
to some narrow application field 
(e.g.~road driving).
Second, we assume that there exists 
a general-purpose dataset
which is much more diverse 
than the inlier dataset (e.g.~ImageNet-1k).
We consider pixels from the general-purpose dataset
as noisy negative samples 
since most (but not all) of them are outliers.
We encourage the model to recognize 
borders between the known and the unknown
by pasting jittered negative patches 
over inlier training images.
Our experiments target two dense 
open-set recognition benchmarks
(WildDash 1 and Fishyscapes) 
and one dense open-set
recognition dataset (StreetHazard).
Extensive performance evaluation indicates
competitive potential of the proposed approach.
\end{abstract}


\end{frontmatter}


\section{Introduction}
\label{s:intro}
  Deep convolutional approaches 
  have recently achieved
  proficiency on 
  realistic semantic segmentation
  datasets such as Vistas \cite{neuhold17iccv}
  or Ade20k \cite{zhou17cvpr}. 
  This success has increased interest 
  in exciting real-world applications
  such as autonomous driving \cite{zhang19pr}
  or medical diagnostics \cite{Xia2020SynthesizeTC}.
  However, visual proficiency of the
  current state-of-the-art models is
  still insufficient to accommodate
  the demanding requirements of
  these applications
  \cite{kendall17nips,nalisnick19iclr}.
  
  Early semantic segmentation approaches
  involved small datasets and few classes.
  Improved methodology and computing power
  led to larger, more diverse datasets
  with more complex taxonomies
  \cite{everingham10ijcv,cordts16cvpr,neuhold17iccv}.
  This development has provided valuable feedback
  that led to the current state of research
  where most of these datasets
  are about to be solved
  in the strongly supervised setup.

  Despite the hard 
  selection and annotation work, 
  most existing datasets are still 
  an insufficient proxy for real-life operation,
  even in a very restricted scenario 
  such as road driving.
  For instance, none of the 20000 images 
  from the Vistas dataset \cite{neuhold17iccv}
  include persons in non-standard poses,
  crashed vehicles or rubble.
  Additionally, real-life images
  may also be degraded due to
  hardware faults, inadequate acquisition,
  or lens distortion \cite{zendel18eccv}.
  This suggests that foreseeing every possible
  situation may be an elusive goal and indicates
  that algorithms should be able to
  recognize image regions 
  foreign to the
  training distribution \cite{kendall17nips}.

  These considerations emphasize 
  the need to further improve 
  the next generation of datasets.
  New datasets should contain 
  atypical images which are likely to fool 
  the current generation of models
  \cite{zendel18eccv,hendrycks2019anomalyseg}.
  Additionally, they should also endorse 
  open-set evaluation \cite{scheirer14}
  where the models are required 
  to perform inference on arbitrary images.
  An open-set model 
  is not supposed to predict 
  an exact visual class in outliers.
  That would often be impossible 
  since the exact visual class 
  may not be present in the training taxonomy. 
  Instead, it should suffice that 
  outliers are recognized as such.
  The desired test subsets should contain 
  various degrees of domain shift 
  with respect to the training distribution.
  This should include diverse contexts
  (e.g.\ adverse weather, exotic locations)
  \cite{neuhold17iccv},
  exceptional situations 
  (e.g.\ accidents, poor visibility)
  \cite{zendel18eccv},
  and outright outliers 
  (foreign domain objects and entire images)
  \cite{zendel18eccv,blum19iccvw}.
  Currently, there are only 
  two such benchmarks
  in the dense prediction domain:
  WildDash \cite{zendel18eccv} and
  Fishyscapes \cite{blum19iccvw}.
  \begin{figure}[h]
   \centering
   \includegraphics[width=\columnwidth]{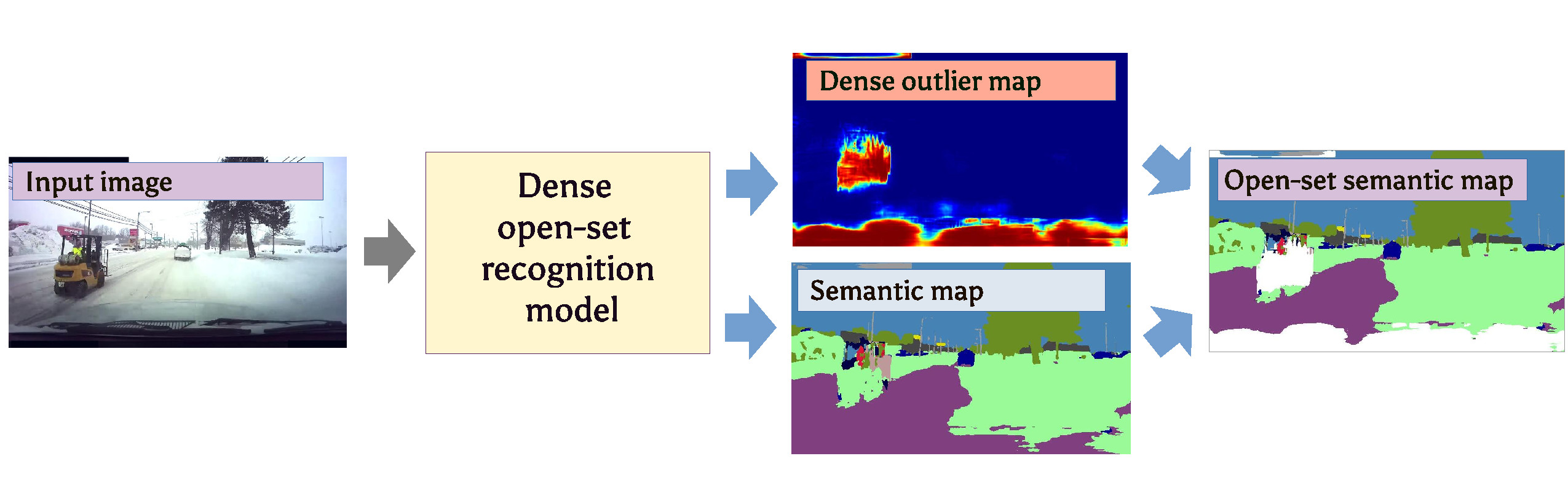}
   \caption{
    A dense open-set recognition model has
    to predict:
    i) a dense outlier map, and 
    ii) a semantic map with C inlier classes. 
    The merged open-set semantic map (right)
    contains outlier pixels (white) 
    on two objects
    which are foreign to the training taxonomy:
    the ego-vehicle and the forklift.
   }
   \label{fig:approach}
  \end{figure}
  
  This paper addresses dense open-set recognition 
  and outlier detection as illustrated in Figure 1.
  Unlike previous \cite{kendall17nips} 
  and concurrent \cite{Xia2020SynthesizeTC} work, 
  we propose to include test-agnostic 
  noisy negatives to the training dataset.  
  We believe that this setup is adequate
  due to extremely large capacity of deep models,
  which allows them to classify outliers into any class
  without hurting empirical accuracy 
  in inliers \cite{zhang17iclr}.
  We believe that our approach will represents
  a strong baseline for some time.
  It is very hard to bound 
  the output of a deep model in foreign samples
  since they may have almost identical latent
  representations to some inliers.
  This holds true
  even for the current state-of-the-art 
  generative models \cite{nalisnick19iclr}.
  
  Our contribution is as follows.
  We propose a novel approach
  for dense outlier detection
  based on discriminative training 
  with noisy negative images 
  from a very large and diverse 
  test-agnostic dataset.
  We show that successful operation
  in dense prediction context
  requires random pasting of negative patches
  to inlier training images.
  Our approach
  can share features with 
  a closed-set semantic segmentation model.
  This greatly improves outlier
  detection while only slightly 
  impairing semantic segmentation.
  Evaluation on two rigorous benchmarks
  and several other datasets indicates 
  that our approach outperforms 
  the state of the art
  \cite{kendall17nips,hendrycks2019anomalyseg,blum19iccvw,Xia2020SynthesizeTC}, 
  especially on large outliers. 
  In datasets with small
  outliers (FS Lost and Found), 
  we achieve the best
  results complementing our
  approach with the max-softmax baseline.
  
  Earlier accounts of this research appeared in 
  \cite{bevandic18arxiv,bevandic19gcpr}. 
  We extend our previous work 
  with improved training procedure,
  broader experimental evaluation, and better results.
  Our consolidated experiments evaluate performance
  on established dense open-set benchmarks
  (WildDash 1 \cite{zendel18eccv}, 
   Fishyscapes Static and Fishyscapes
   Lost and Found \cite{blum19iccvw}),
   the StreetHazard dataset \cite{hendrycks2019anomalyseg},
   and the proposed WD-Pascal dataset
   \cite{bevandic18arxiv,bevandic19gcpr}. 
  Our experiments show that
  the proposed approach 
  is broadly applicable
  without any dataset-specific tweaking.
  All our experiments 
  use the same negative dataset
  and involve the same hyper-parameters.
  The resulting models produce
  dense open-set prediction
  with a single forward pass,
  which makes them suitable
  for real-time inference.
  
\section{Related work}
Open-set recognition combines
classification and outlier detection.
Some novelty detection approaches
can be viewed as open-set recognition
though this connection is seldom discussed.
We are especially concerned with
dense open-set recognition
and focus on approaches
that train on negative data.

\subsection{Open-set recognition}
Open-set recognition involves
C known classes during training
and (C+1) classes during inference.
The (C+1)st label signifies
that the sample does not belong
to the training distribution.
Outliers are usually recognized by
thresholding some kind of score.

Open-set classification can be formulated
on top of a classic
closed-set discriminative model
by estimating the outlier score
from the prediction itself.
Most recent work considers
the probability of the winning class,
also known as max-softmax (MSM)
\cite{hendrycks17iclr}.
Unfortunately, deep models usually
have highly confident outputs
regardless of the input
\cite{guo17icml}.
Different strategies
can make max-softmax more informative,
e.g.\ recalibration \cite{guo17icml},
preprocessing \cite{liang18iclr},
MC-Dropout \cite{gal16icml}
or ensembling
\cite{Bergmann_2020_CVPR}.
However, recalibration cannot improve
average precision (AP).
Preprocessing and MC-dropout offer
only slight improvements over the baseline.
MC-Dropout and ensembling require
multiple forward passes,
which may not be acceptable
for large images
and real-time inference.

Prediction uncertainty can also be assessed
with a jointly trained head
of the compound model.
The two heads operate on shared features
for efficiency and cross-task synergy
\cite{devries18arxiv, kendall17nips, zhang2020hybrid}.
Unfortunately, this can only recognize
aleatoric uncertainty \cite{kendall17nips}
which may arise due to inconsistent labels.
Instead, outlier detection is related
with epistemic uncertainty
which arises due to insufficient learning
\cite{kendall17nips,hullermeier21ml}.
Epistemic uncertainty has been assessed
under assumption that MC dropout
approximates Bayesian model sampling
\cite{smith18uai}. However, that
assumption may not be satisfied in practice.
Additionally, existing approaches
\cite{kendall17nips,smith18uai}
confound model uncertainty
with distributional uncertainty
\cite{malinin18nips}.

We are especially interested
in approaches which exploit
negative samples during training.
Most of these approaches complement
the standard discriminative loss
with a term which encourages
high entropy in negative samples,
such as KL-divergence towards a suitable prior
\cite{lee18iclr, hendrycks19iclr,malinin18nips}.
A negative dataset can also be exploited
to train a separate prediction head
which directly predicts the outlier probability
\cite{bevandic18arxiv}.
However, these approaches are sensitive
to the choice of the negative dataset.
An alternative approach
trains on synthetic negatives
which are generated at the border
of the training distribution.
However, experiments suggest
that diverse negative datasets
lead to better outlier detection
than synthetic negative samples
\cite{lee18iclr,hendrycks19iclr}.

\subsection{Novelty detection}

Novelty detection is an umbrella term 
which covers anomaly, rare-event, 
outlier and OOD detection, and
one-class classification.
Most of this work addresses generative models
which attempt to model
the training distribution.
Anomalous examples should yield
low probabilities in this setup, 
though this is 
difficult to achieve in practice
\cite{nalisnick19iclr,GrathwohlWJD0S20}.
Generative adversarial networks 
can be used to score 
the difference between the input
and the corresponding reconstruction \cite{zenati18}
if the generator is formulated
as an auto-encoder where the
latent representation mapping
is trained simultaneously alongside
the GAN \cite{ZHANG2020PR}.
However, the obtained reconstructions 
are usually imperfect 
regardless of the type of input
\cite{Bergmann_2020_CVPR}.

Several works emphasize contribution of
knowledge transfer \cite{Bergmann_2020_CVPR}, 
although fine-tuning gradually
diminishes pre-training benefits
due to 
forgetting. This effect can 
be somewhat attenuated
with a modified loss \cite{perera19}.

\subsection{Dense open-set recognition}

Dense open-set recognition is still an 
under-researched field
despite important applications 
in intelligent transportation \cite{zhang19pr}
and medical image analysis \cite{Xia2020SynthesizeTC}.
Some of the described novelty detection methods 
are capable of dense inference
\cite{Bergmann_2020_CVPR},
however they address simple datasets 
and do not report pixel-level metrics.
Hence, it is unclear whether they could be 
efficiently incorporated into 
competitive semantic segmentation frameworks.

Many image-wide open-set approaches
can be adapted for dense prediction 
straightforwardly
\cite{kendall17nips,bevandic19gcpr,blum19iccvw}
though they are unable
to achieve competitive performance
due to many false positive outlier detections.
This likely occurs 
because dense prediction incurs 
more aleatoric uncertainty 
than image-wide prediction 
due to being ill-posed 
at semantic borders \cite{bevandic18arxiv}.

A concurrent approach \cite{blum19iccvw}
fits an ensemble of normalized flows
to latent features of the segmentation model.
They infer negative log likelihoods
in different layers and threshold with respect   
to the most likely activation across all layers.
This approach achieves a fair accuracy 
on the Fishyscapes benchmark,
however our submission outperforms it.

Preceding discussions suggest
that dense open-set recognition
is a challenging problem,
and that best results may not be attainable
by only looking at inliers.
Our work is related to two recent 
image-wide outlier detection approaches
which leverage negative data.
Perera et al.~\cite{perera19}
learn features for one-class classification
by simultaneously optimizing
cross-entropy on ImageNet images
and feature compactness on the target images.
However, inlier compactness and template-matching 
are not suitable for complex training ontologies.
Hendrycks et al.~\cite{hendrycks19iclr}
train a discriminative classifier 
to output low confidence in negative images.
However, our experiments suggest tendency
towards false positives 
due to aleatoric uncertainty at semantic borders.

Our work is also related to the
dense open-set recognition approach
which treats outlier detection 
by extending the inlier ontology \cite{MSeg_2020_CVPR}.
The proposed composite dataset (MSeg)
collects almost 200\,000 
densely annotated training images
by merging public datasets
such as Ade20k, IDD, COCO etc.
Currently, this is the only approach 
that outperforms our submission 
to the WildDash 1 benchmark.
However, the difference in performance 
is only 1.4pp although we train on 
smaller resolution (768 vs 1024) and
use less negative supervision during training
(bounding boxes from ImageNet-1k instead of
 dense labels on COCO, Ade20k and SUN RGB-D).
Their approach does not appear 
on the Fishyscapes leaderboard.

\section{Method}
\label{ss:model}
\begin{figure}[hbt]
\centering
\includegraphics[width=\columnwidth]{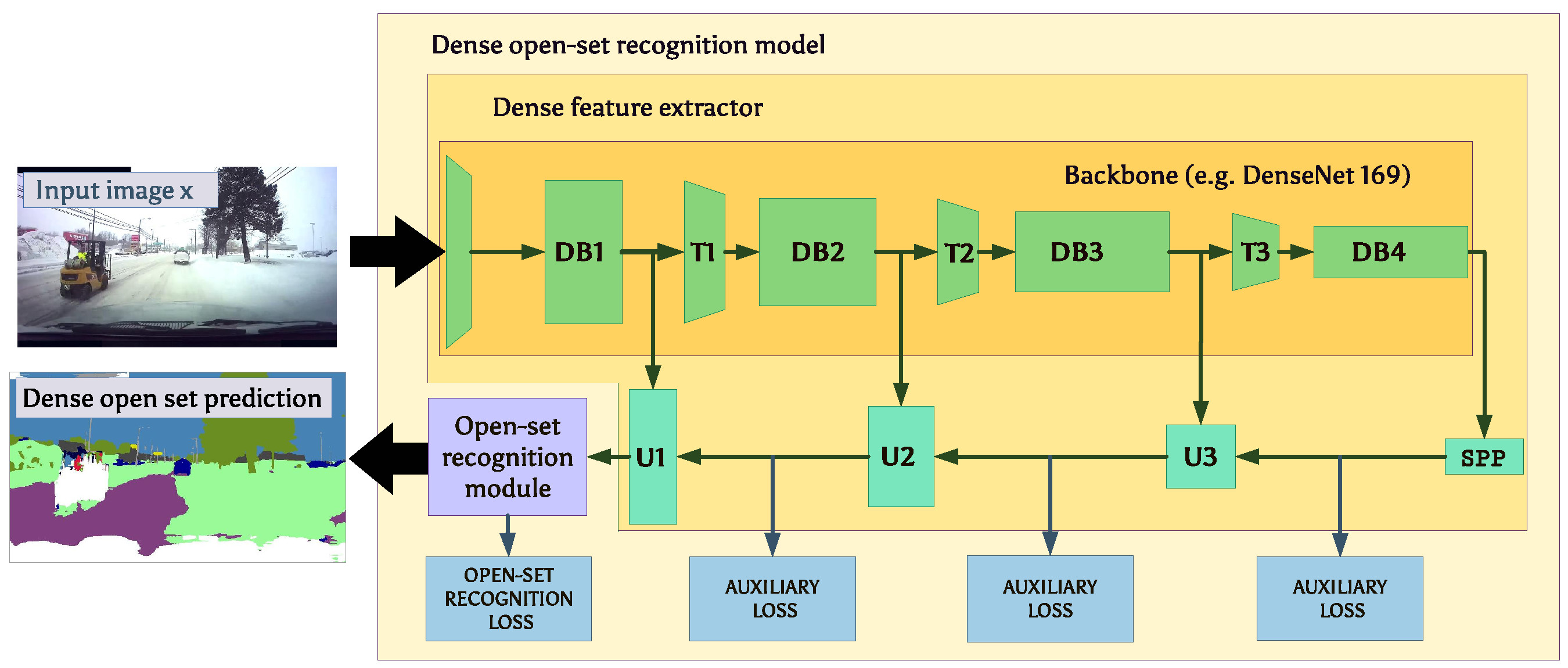}
\caption{
  The proposed dense open-set recognition model
  consists of a dense feature extractor
  and a dense open-set recognition module.
  The dense feature extractor contains densely
  connected blocks (DB), 
  transition blocks (T), spatial
  pyramid pooling layer (SPP) and
  lightweight upsampling blocks (U)
  \cite{kreso20tits}.
  We use auxiliary cross-entropy
  losses to speed-up and regularize
  training.
  The open-set recognition module 
  produces semantic segmentation into C+1
  classes, where the C+1st
  class is the outlier class.
}
\label{fig_sim}
\end{figure}

The main components of our approach
are the dense feature extractor
and the open-set recognition module
illustrated in Fig.\,\ref{fig_sim}.
The dense feature extractor
is a fully convolutional module
which transforms the input image 
H$\times$W$\times$3 into
a shared abstract representation
H/4$\times$W/4$\times$D 
where D is typically 256.
The dense open-set recognition module
incorporates recognition and outlier detection.
We base these two tasks on shared features 
to promote fast inference
and cross-task synergy 
\cite{devries18arxiv,perera19}.
Our method relies on the following two hypotheses:
i) training with diverse noisy negatives 
   can improve outlier detection 
   across various datasets, and 
ii) shared features greaatly improve
outlier detection without significant
deterioration of semantic segmentation.

\subsection{Dense feature extraction}
\label{ssec:featex}
Our feature extraction module consists of
a powerful downsampling path 
responsible for semantics,
and a lean upsampling path
which restores the spatial detail.
The downsampling path starts with a 
pre-trained recognition backbone.
In case of DenseNet-169 it consists of
four densely connected blocks (DB1-DB4) and 
three transition layers (T1-T3). Lightweight 
spatial pyramid pooling (SPP)
provides wide context information
\cite{kreso20tits, zhao17cvpr}.
The upsampling path consists of 
three upsampling modules (U1-U3)
which blend low resolution features
from the previous upsampling stage
with high-resolution features 
from the downsampling path.
The resulting encoder-decoder structure is asymmetric.
It has dozens of convolutional layers 
in the downsampling path
and only three convolutional layers 
along the upsampling path
\cite{kreso17cvrsuad}.
We speed-up and regularize the learning
with auxiliary cross-entropy losses. 
These losses target soft
ground truth distribution 
across the corresponding window
at full resolution \cite{kreso20tits}.

\subsection{Two-head recognition module}
\label{ss:two-head-module}
We consider dense open-set recognition  
with shared features.
We assume that the training data $\mathcal{D}$
contains both inlier
and noisy negative pixels.
We denote images with $\mathbf{x}$,
dense semantic predictions with $\mathbf{Y}$ 
and the corresponding 
C-way ground truth labels with $\mathbf{y}$.
Similarly, dense outlier predictions
and the corresponding ground truth labels 
are $\mathbf{O}$ and $\mathbf{o}$,
respectively. We use $i$ and $j$
to denote the location of pixels.
Most considerations become applicable
to image classification 
by removing summation over all pixels (i,j)
and regarding $Y_{ij}$ and $O_{ij}$ 
as image-wide predictions.

We propose a two-head open set 
recognition module which simultaneously emits
dense closed-set posterior over classes
$\mathrm{P}(Y_{ij}|\mathbf{x})$, as well as
the probability $\mathrm{P}(O_{ij}|\mathbf{x})$
that the pixel at coordinates $(i,j)$ is an outlier.
Standard cross-entropy losses
for the two predictions are as follows:
\begin{myalign}
    \label{eq:two_head}
  \mathcal{L}_\mathrm{cls} &= 
    -
    \sum_{\mathbf{x},\mathbf{y}, \mathbf{o}\in
           \mathcal{D}} 
     \sum_{ij} 
      [\![o_{ij}=0]\!] \cdot \log \mathrm{P}(Y_{ij}=y_{ij}|\mathbf{x}) 
    \; ,
    \nonumber \\ 
  \mathcal{L}_\mathrm{od} &= 
    -
    \sum_{\mathbf{x,o}\in
      \mathcal{D}} 
    \sum_{ij}
      \log \mathrm{P}(O_{ij}=o_{ij}|\mathbf{x})
    \; .
\end{myalign}

\begin{figure}[htb]
\centering
\includegraphics[width=0.95\linewidth]{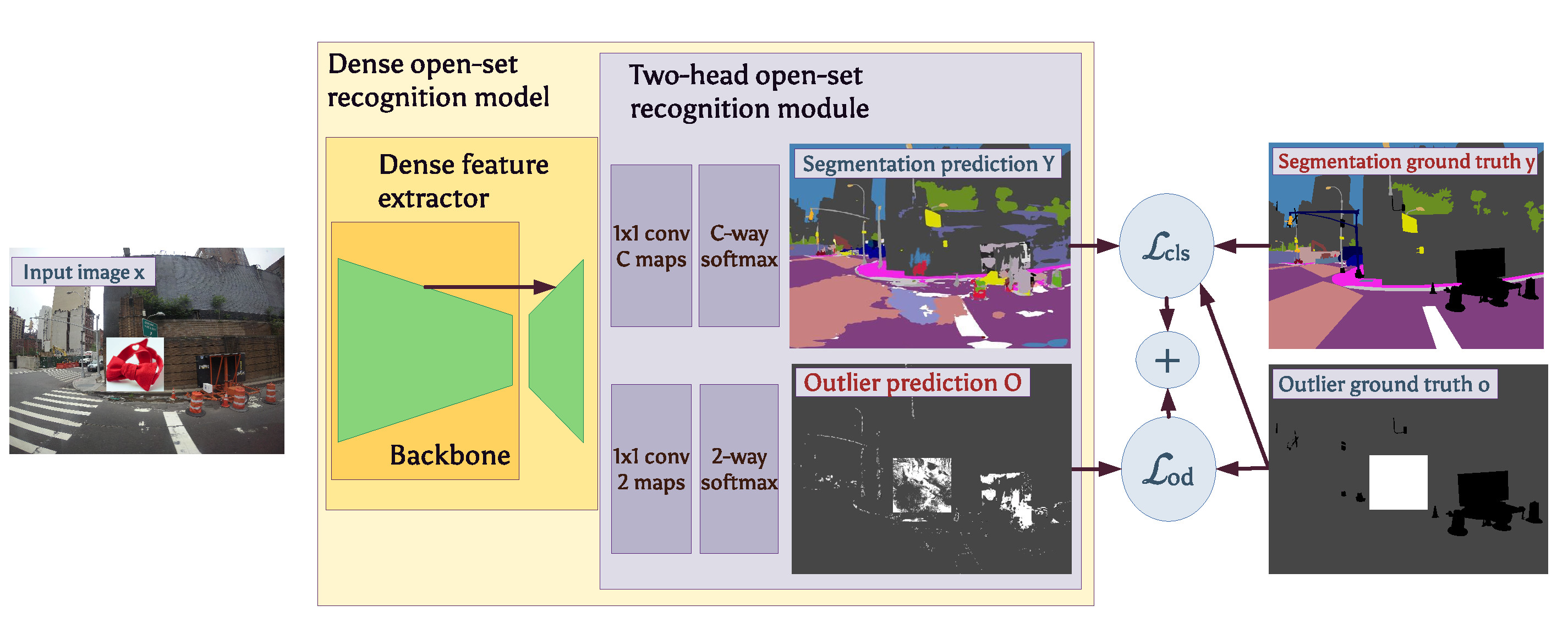}
\caption{The architecture of the proposed
two head open-set recognition module.
The outlier detection head is a binary 
classifier which we train using the 
outlier ground truth.
The segmentation head is a C-way classifier
which requires both the segmentation 
and the outlier ground truth. The
outlier ground truth is required for segmentation 
training in order to be able to
exclude outlier pixels from
$\mathcal{L}_\mathrm{cls}$.
}
\label{fig:two-head-module}
\end{figure}

Figure
\ref{fig:two-head-module} shows 
that equation \ref{eq:two_head} can be
implemented as a multi-task model with shared
features where the first
head predicts semantic segmentation,
while the second detects outliers.

Outlier detection overrides closed-set recognition
when the outlier probability is over a threshold.
Thus, the classification head 
is unaffected by negative data,
which may preserve 
the baseline recognition accuracy
even when training on test-agnostic negatives
which are bound to be noisy.

\subsection{Exploiting noisy negatives}
\label{ssec:noisy_negatives}
We propose to train our model
by sampling negative data 
from an extremely diverse test-agnostic dataset
such as ImageNet-1k.
We observe that such dataset
will necessarily
overlap with inliers.
For example, ImageNet-1k
contains many classes 
from road-driving ontologies
used in Cityscapes \cite{cordts16cvpr}
and Vistas \cite{neuhold17iccv}
(e.g.\ cab, streetcar).
Additionally, most stuff 
classes from Cityscapes
(e.g.\ building, vegetation) 
are a regular occurrence 
in ImageNet-1k backgrounds.
We refer to this issue as label noise.

We promote resistance to label noise 
by training on mixed batches 
with approximately equal share 
of inlier and negative images.
Hence, inlier pixels in negative images
are vastly outnumbered by true
inliers for each particular class.
We perform many inlier epochs
during one negative epoch,
since our negative training dataset 
is much larger than our inlier datasets.
Our batch formation procedure
prevents occasional inliers from negative images
to significantly affect the training
and favours stable development 
of batchnorm statistics.
Unlike \cite{blum19iccvw},
we refrain from training on 
pixels labeled with the ignore class
since we wish to use the same
negative dataset in all
experiments.

\begin{figure}[hbt]
\begin{subfigure}{0.5\textwidth}
\centering
\includegraphics[width=0.95\linewidth]{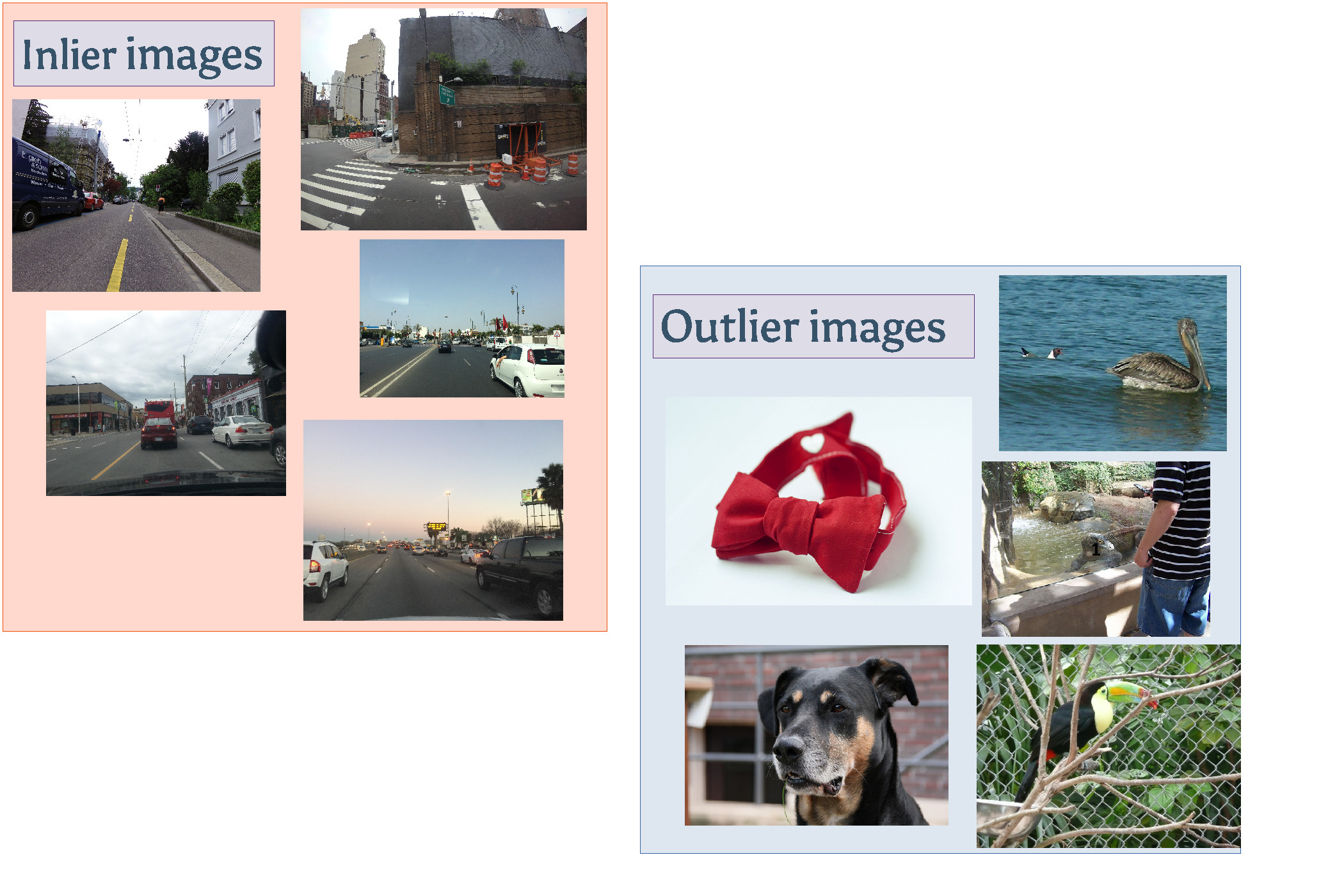}
\caption{ \\ }
\label{fig:neg}
\end{subfigure}
\begin{subfigure}{0.5\textwidth}
\centering
\includegraphics[width=0.95\columnwidth]{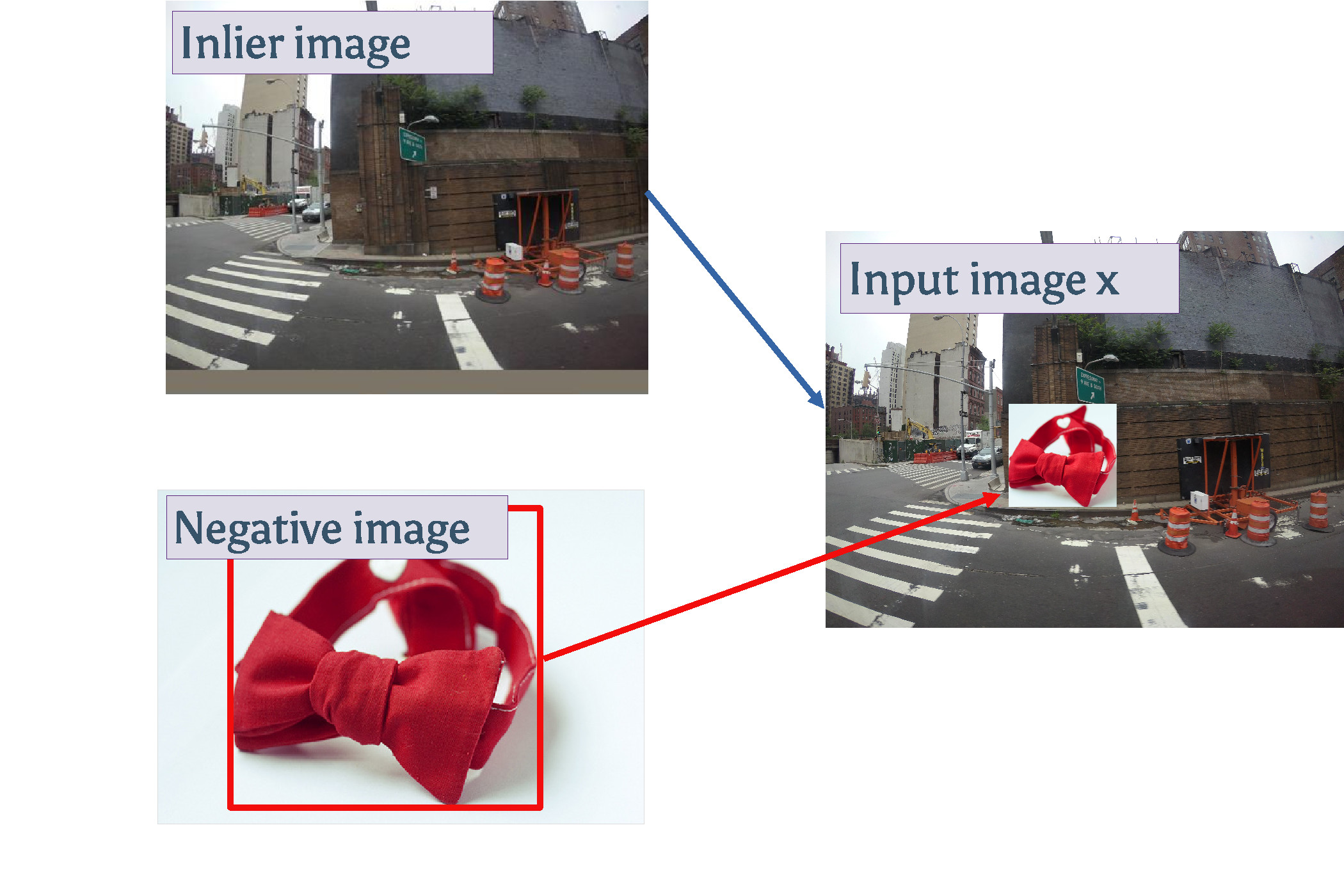}
\caption{ \\ }
\label{fig:neg_gt}
\end{subfigure}
\\
\bigskip
\begin{subfigure}{0.5\textwidth}
\centering
\includegraphics[width=0.95\columnwidth]{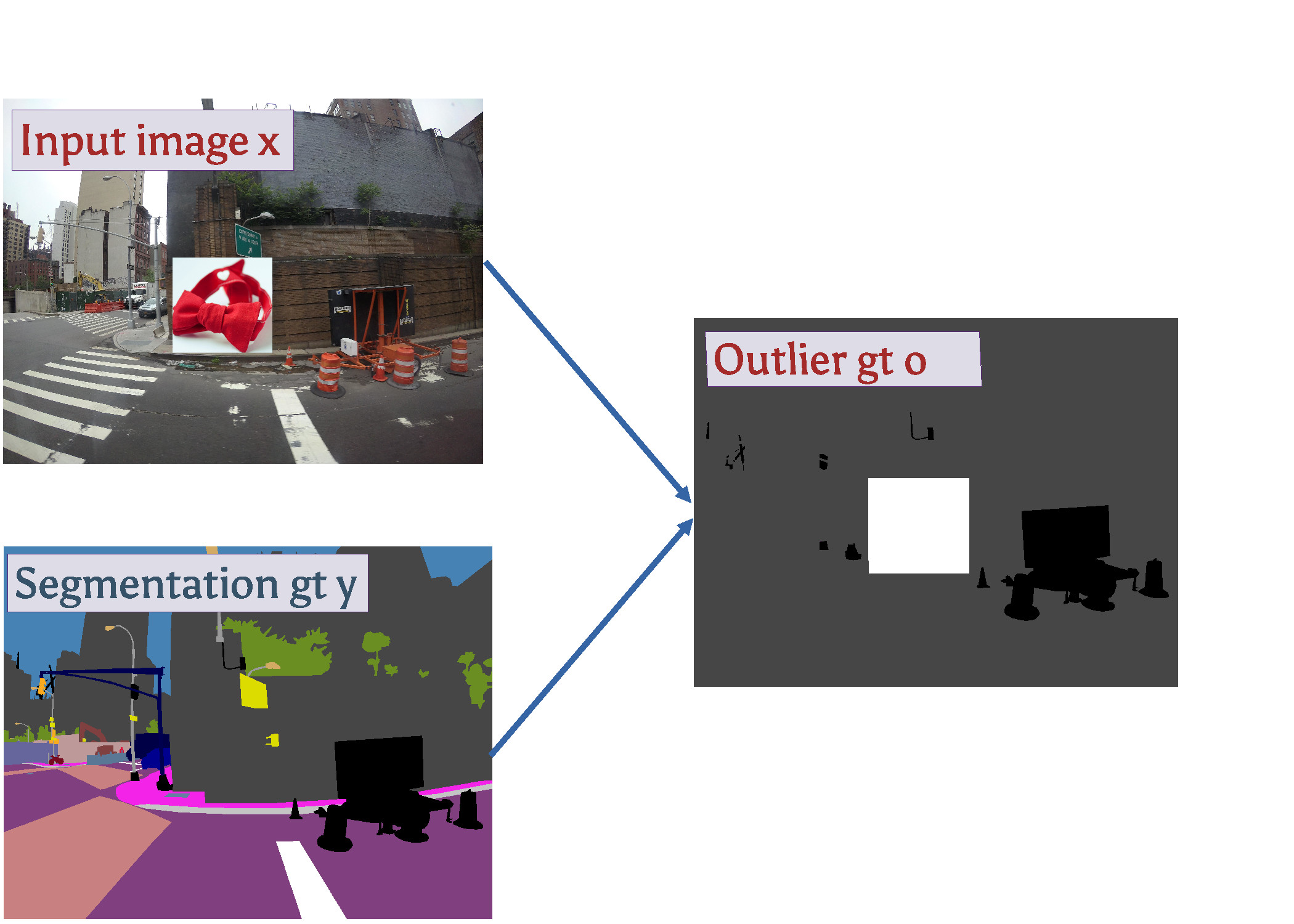}
\caption{ }
\label{fig:pasted}
\end{subfigure}
\begin{subfigure}{0.5\textwidth}
\centering
\includegraphics[width=0.95\linewidth]{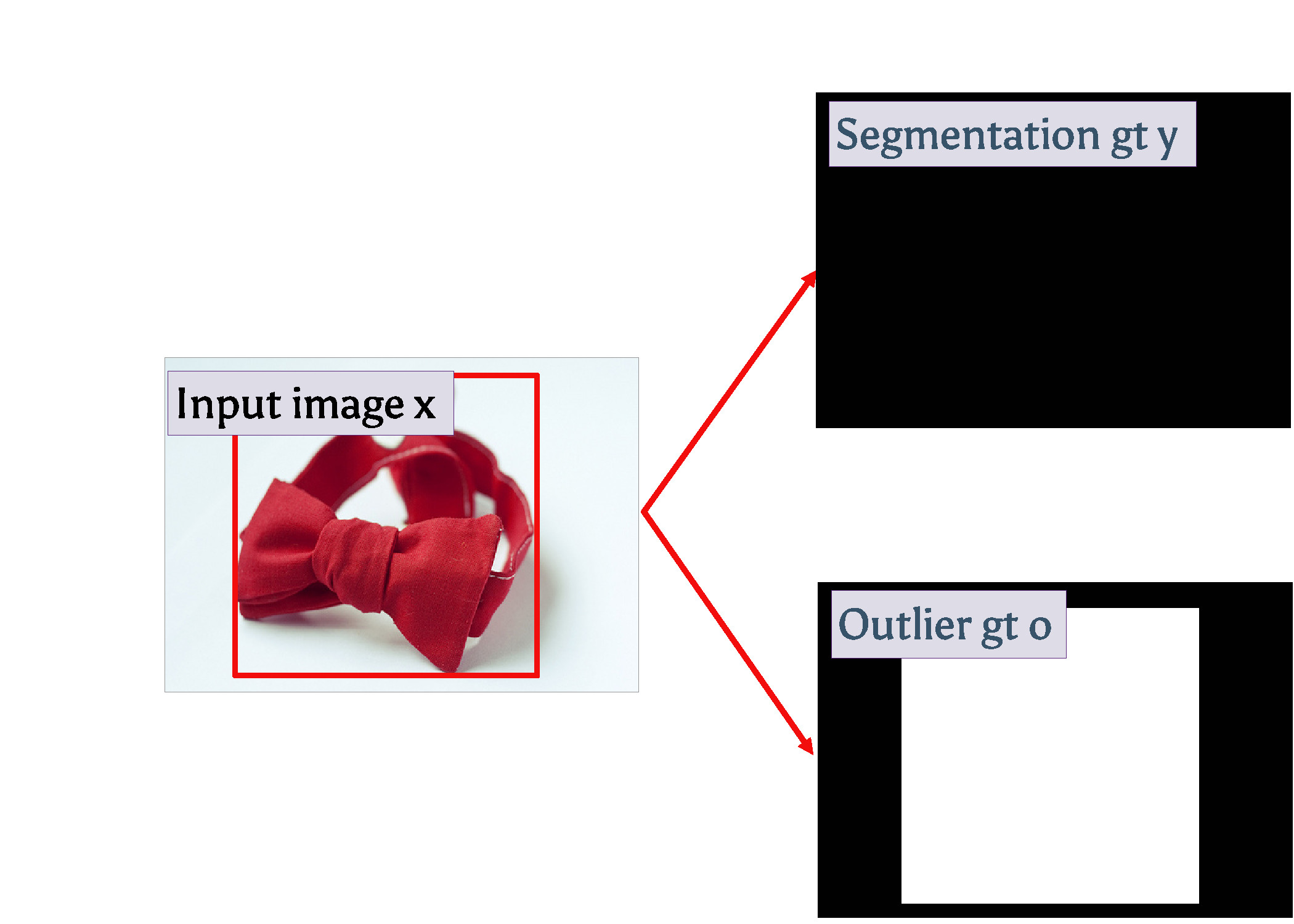}
\caption{ }
\label{fig:pasted_gt}
\end{subfigure}
\caption{We train on images from the target dataset and noisy
negatives from ImageNet-1k (a). We paste a randomly rescaled
noisy negative bounding box into each positive training image
(b). The pasted pixels are labeled as outliers (white) in the
outlier detection ground truth (c). Negative training images are
completely ignored by the semantic segmentation loss (black) and
labeled as outliers only within the bounding box (d).}
\label{fig:training_input}
\end{figure}

Our early experiments involved
training on whole inlier images
and whole negative images.
The resulting models would work very well
on test images with all inliers or all outliers.
However, the performance was poor in images 
with mixed content \cite{bevandic18arxiv}.
It appears that the outlier detection head
must be explicitly trained for mixed inputs
to correctly generalize in such cases.
We address this issue by pasting negative images 
into inlier images during training.
We first resize the negative image 
to a small percent of the inlier resolution,
and then paste it at random in the inlier image
as illustrated in Figure \ref{fig:training_input}.
Subsequently, our models became capable 
of detecting outliers in inlier context
\cite{bevandic19gcpr}.
We obtain the best results 
when the size of pasted patches
is randomly chosen from a wide interval.

\section{Experimental setup}
Our open-set recognition models
aim at achieving robustness
with respect to various forms
of distributional uncertainty
\cite{malinin18nips}.
Consequently all our experiments
evaluate on datasets
which are in some way
different than the training ones.

\subsection{Training datasets}
We train our models on inliers
from Cityscapes train \cite{cordts16cvpr},
Vistas train \cite{neuhold17iccv},
and StreetHazard train \cite{hendrycks2019anomalyseg}.
We train all our models on the same
noisy negative training dataset
which we refer to as ImageNet-1k-bb
\cite{bevandic19gcpr}.
We collect ImageNet-1k-bb by picking
the first bounding box
from the 544546 ImageNet-1k images
with bounding box annotations.
We train on standalone negative images
and mixed-content images obtained by pasting
a resized negative image into an inlier crop.
We resize each negative image
to the desired share $s_n$
of the inlier resolution,
where the default is $s_n$=5\%.
Models with the RSP suffix
(randomly scaled patches)
pick a random $s_n\in[.1\%,10\%]$
for each negative training image.

\subsection{Validation dataset}
Several previous approaches propose
to evaluate dense open-set recognition
on splits of existing real datasets
that contain some visual classes
which are absent from the training split.
Thus, the BDD-Anomaly dataset
\cite{hendrycks2019anomalyseg}
collects all BDD images
without trains and motorcycles
into the training split,
and places all other BDD images
into the test split.
Cityscapes-IDD \cite{angus19arxiv}
proposes training on Cityscapes,
and evaluating on cars (inliers)
and rickshaws (outliers)
from the IDD dataset.
However, this approach
is not easily carried out in practice
since it is hard to avoid similarities
between inlier and outlier classes.
For instance, trains and motorcycles
are similar to buses and bicycles,
respectively, which are inliers in BDD-Anomaly.
Similarly, rickshaws (Cityscapes-IDD outliers)
are similar to motorcycles and cars
(Cityscapes-IDD inliers).
We attempt to avoid this pitfall
by making sure that anomalies
come from a different domain.
We craft WD-Pascal \cite{bevandic18arxiv}
by randomly pasting Pascal animals
into WildDash 1 val images.
We select animals which take up
at least 1\% of the WildDash resolution.
Conversely, we craft WD-LSUN
by complementing WildDash 1 val
with random subsets
of LSUN \cite{yu15arxiv} images,
so that the number of inliers (WildDash 1)
and outliers (LSUN) is approximately equal.
We reduce the variance of all our
validation and ablation experiments
by averaging 50 assays across WildDash 1.

\subsection{Evaluation datasets}
We evaluate our models on
several test dataset
for dense open-set recognition.
Our experiments report
the outlier detection performance
(AP, FPR$_{95}$ \cite{hendrycks17iclr})
and semantic segmentation accuracy (mIoU).
The WildDash 1 benchmark \cite{zendel18eccv}
collects difficult road driving scenarios
and negative images from other domains,
but does not include images of mixed content.
The Fishyscapes benchmark \cite{blum19iccvw}
includes Cityscapes images
with pasted Pascal VOC objects.
It also includes a subset of the
Lost and Found dataset \cite{pinggera16iros}
where the outliers correspond
to small obstacles on the road.
The StreetHazards dataset
\cite{hendrycks2019anomalyseg}
contains fully synthetic road-driving images
while out-of-domain objects
correspond to anomalies.

\subsection{Implementation details}
Our models are based on DenseNet-169
with ladder-style upsampling \cite{kreso20tits}
as described in \ref{ssec:featex}
due to best overall validation
performance \cite{bevandic19gcpr}.
We normalize all images
with ImageNet mean and variance.
We denote the image size
as its shorter dimension.
We resize WD-Pascal and
WD-LSUN images to 512 pixels.
In all other experiments we resize
validation and test images to 768 pixels.
Some experiments train with scale jittering
so that 30\% images are resized to 512 pixels,
while the remaining 70\% images
are randomly resized
between 512 and 1536 pixels.
We denote such models with
the JS suffix (jittered scale).
We form training batches
with random 512$\times$512 crops
which we jitter with horizontal flipping.
We do not use multi scale evaluation
in order to report performance
which could be delivered in real-time.We use the standard Adam optimizer
and divide the learning rate
of pre-trained parameters by 4.
We validate the loss weights of all
open-set recognition modules on a
small subset of WD-Pascal.
We train our two-head models with the compound loss 
$\mathcal{L}_{th} = 0.6 \mathcal{L}_{cls} + 0.6*0.2 \mathcal{L}_{od} + 0.4 \mathcal{L}_{aux}$.
We validate all hyper-parameters
on WD-Pascal and WD-LSUN
\cite{bevandic19gcpr}.
We train our models throughout 75 Vistas epochs,
which corresponds to 5 epochs of ImageNet-1k-bb.
This was increased to 20 epochs
for our benchmark submissions.
We detect outliers
by thresholding inlier probability
at $\mathrm{P}(O_{ij}=0|\mathbf{x})$=0.5.

\section{Results}

We validate mIoU accuracy on WildDash 1 val
and outlier detection AP on
WD-Pascal, WD-LSUN and 
Fishyscapes Lost and Found.
We evaluate our models on
the WildDash 1 benchmark,
the Fishyscapes benchmark, 
and on the test subset of the 
StreetHazard dataset.

\subsection{Validation of Dense 
  Outlier Detection Approaches}

\begin{figure}[htb]
\begin{subfigure}{0.5\textwidth}
\centering
\includegraphics[width=0.95\columnwidth]{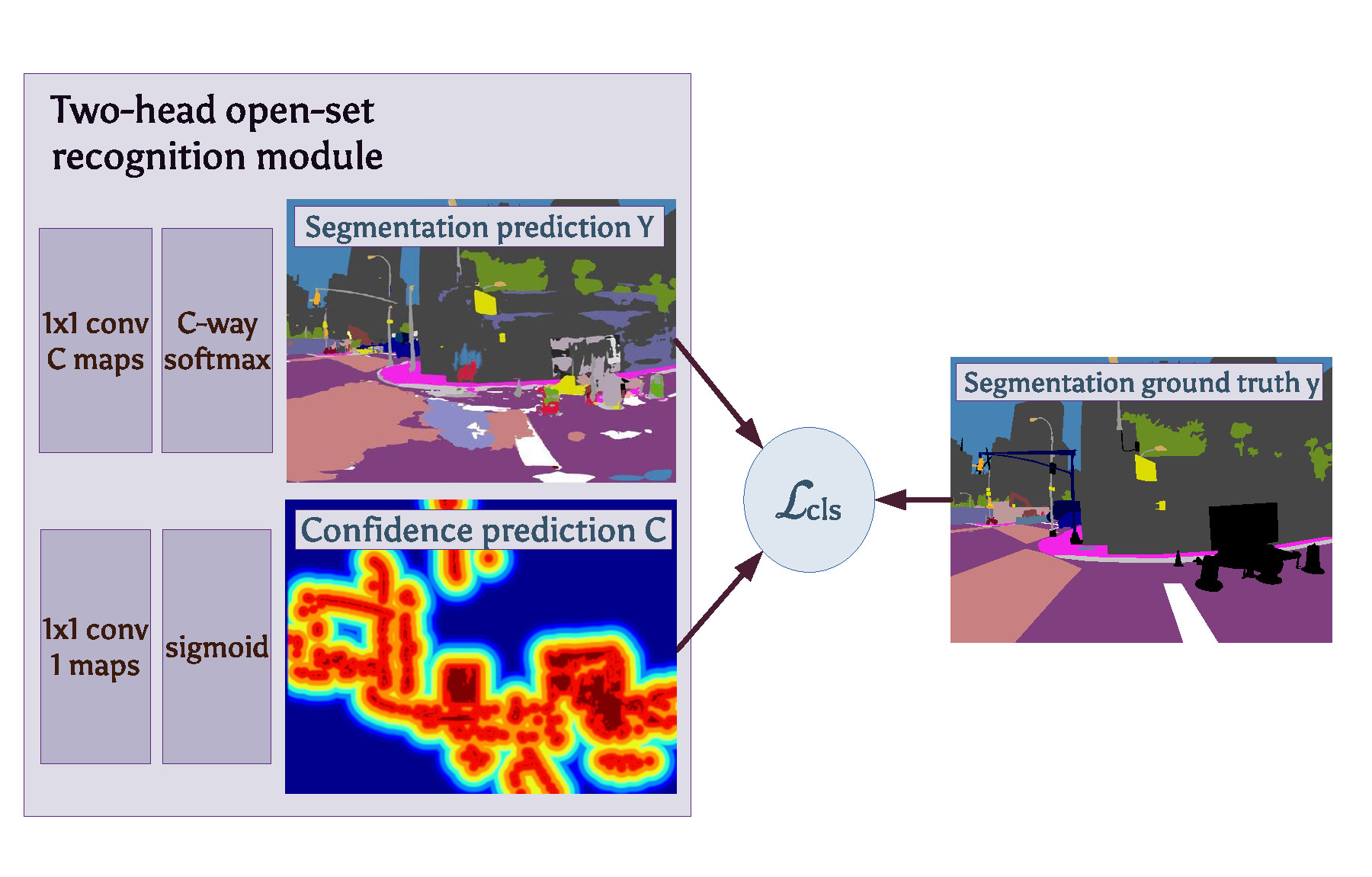}
\caption{ }
\label{fig:conf-module}
\end{subfigure}
\begin{subfigure}{0.5\textwidth}
\centering
\includegraphics[width=0.95\linewidth]{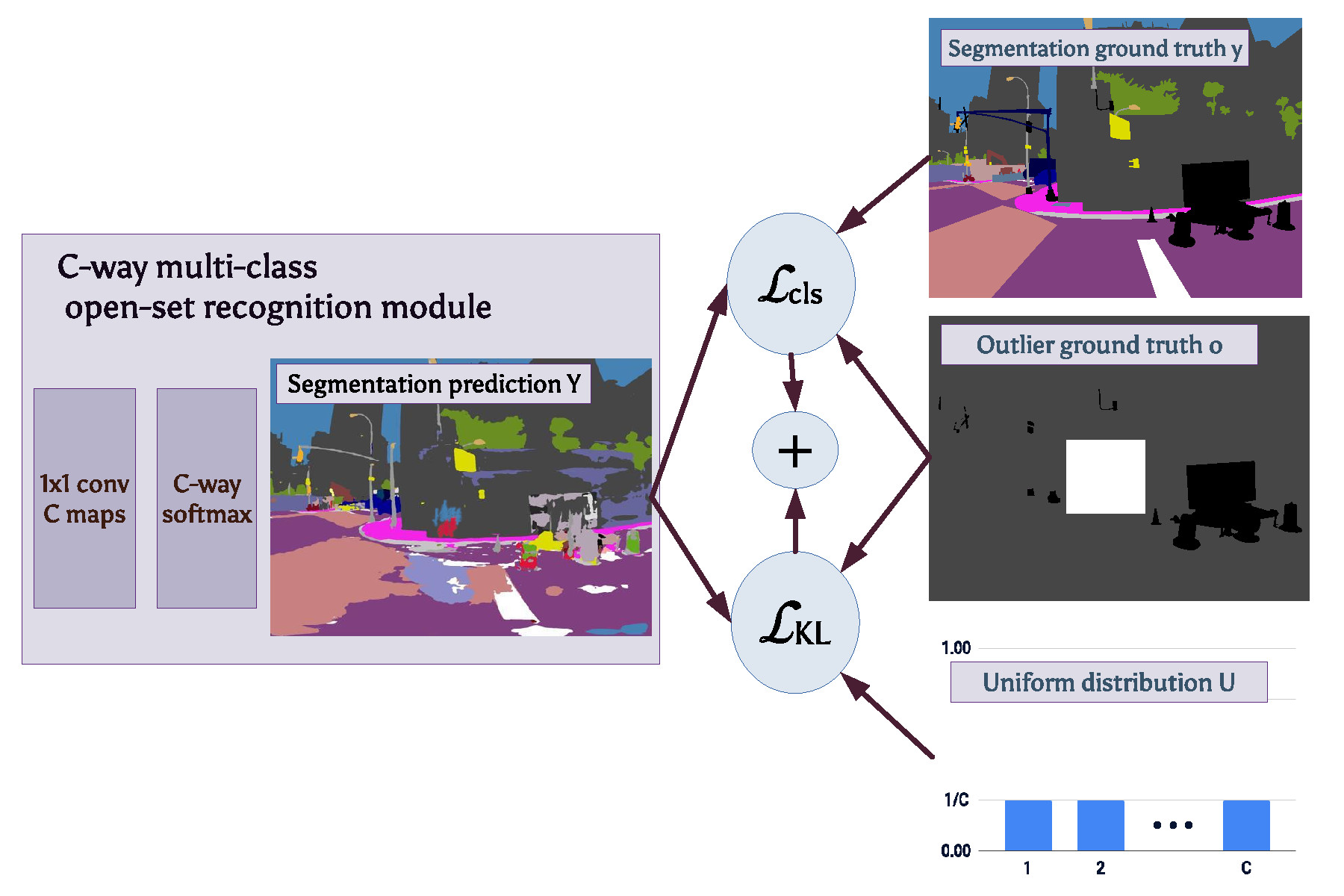}
\caption{ }
\label{fig:oe_module}
\end{subfigure}
\\
\begin{subfigure}{0.5\textwidth}
\centering
\includegraphics[width=0.95\columnwidth]{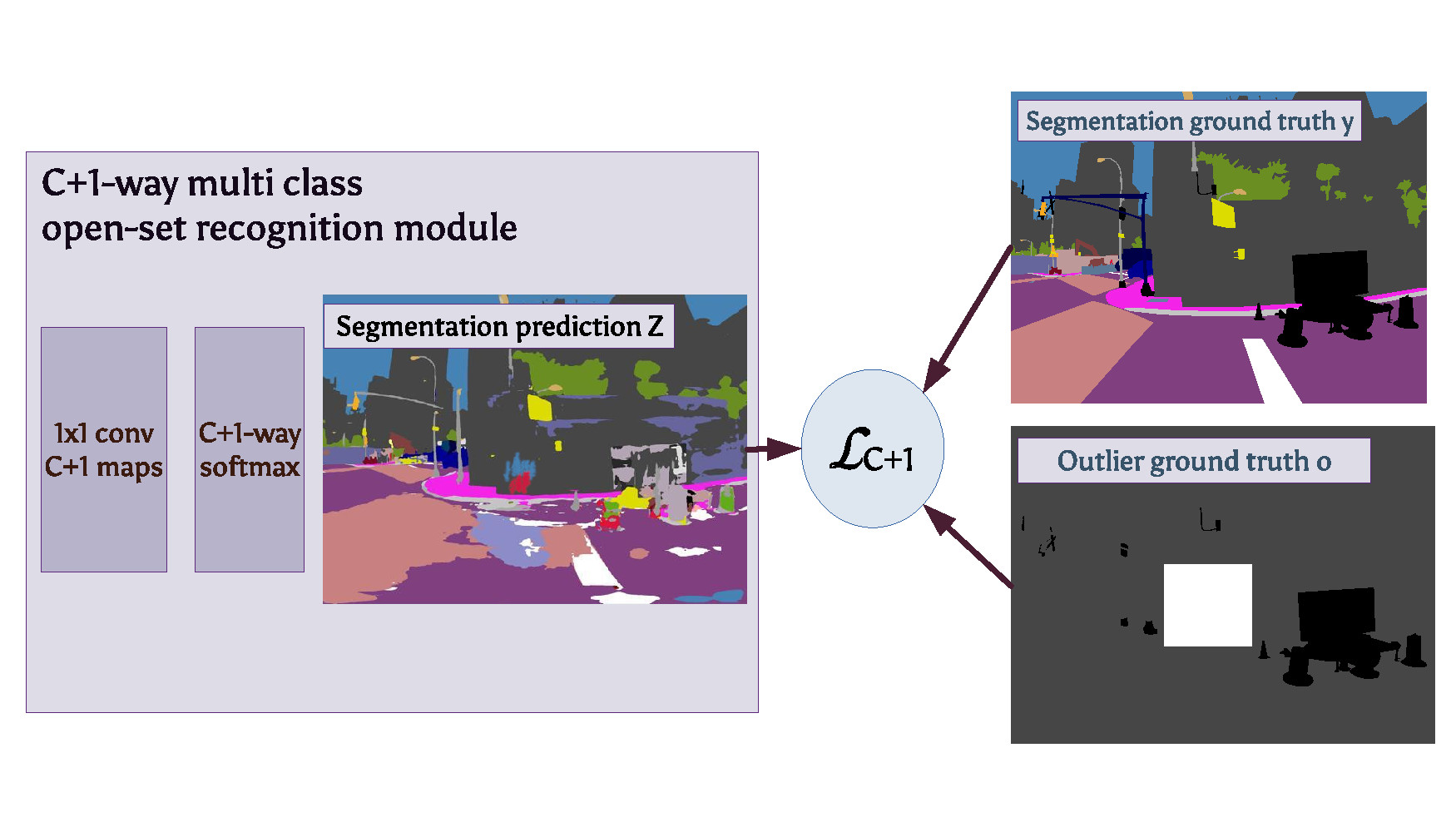}
\caption{ }
\label{fig:c+1-module}
\end{subfigure}
\begin{subfigure}{0.5\textwidth}
\centering
\includegraphics[width=0.95\columnwidth]{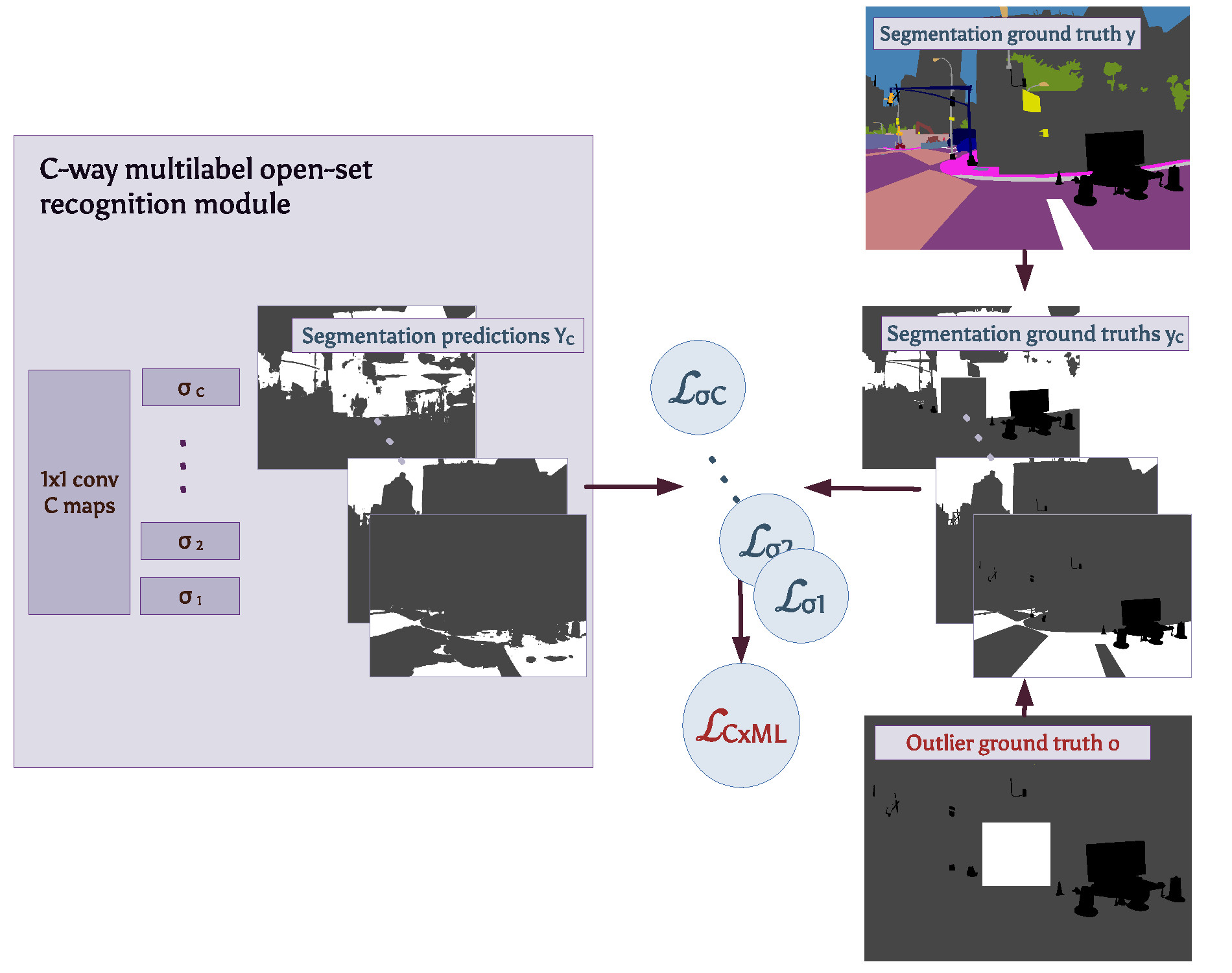}
\caption{ }
\label{fig:sigmoid_module}
\end{subfigure}

\caption{Four alternative open-set recognition modules.
Two-head approach with trained confidence
\cite{devries18arxiv, kendall17nips} is
similar to our approach in Figure 3,
but it does not train on negative images (a).
C-way multi-class approach \cite{hendrycks19iclr, lee18iclr}
learns uniform prediction in negative samples (b).
C+1-way multi-class approach uses the
negative data as a regular semantic class (c).
C-way multi-label approach learns C one-versus-all
classifiers \cite{franchi20arxiv} (d).}
\label{fig:dense_osr_modules}
\end{figure}

Table \ref{table:OOD_detection} validates
our method against several other dense
open-set recognition approaches 
on WD-Pascal and WD-LSUN. 
All models have been trained on positive images
from the Vistas dataset.
Section 1 of the table presents
models which are trained without negatives.
We show the performance of max-softmax \cite{hendrycks17iclr},
max-softmax after ODIN \cite{liang18iclr}
epistemic uncertainty after 50 forward
passes with MC-Dropout \cite{smith18uai}, 
and densely trained confidence \cite{devries18arxiv}
(cf. Figure \ref{fig:conf-module}).

The remaining models use noisy
negatives from ImageNet-1k-bb during training.
Section 2 of the table
evaluates a single-task outlier detection
model.
The model performs better than
the models from section 1, but much worse than
models from section 4 which share features
between the segmentation and the outlier 
detection tasks.
This confirms our hypothesis that semantic segmentation loss
forces the model to learn features
which generalize well for outlier detection.

Section 3 evaluates the two-head module
approach from Figure \ref{fig:two-head-module}
when it is trained on whole inlier and whole
negative images. This model is able to 
detect outlier images but it performs
badly on images with mixed content.
This shows that training with 
pasted negatives is a prerequisite
for detecting outlier objects in
front of an inlier backgroung.

\begin{table}[htb]
\centering
\resizebox{\textwidth}{!}{
\begin{tabular}{lrrr}
  Model &
    \multicolumn{1}{r}{AP WD-LSUN} &
    \multicolumn{1}{r}{AP WD-Pascal} &
    \multicolumn{1}{r}{mIoU WD}\\
    \toprule
  \multicolumn{1}{l}{
    C$\times$ multi-class}  &$55.6 \pm 0.8$ & $5.0 \pm 0.5$ & 50.6
    \\
  \multicolumn{1}{l}{
    C$\times$ multi-class, ODIN}  & $56.0 \pm 0.8$ & $6.0 \pm 0.5$ & \textbf{51.4} 
    \\

  \multicolumn{1}{l}{
    C$\times$ multi-class, MC}  & $\textbf{64.1} \pm \textbf{1.0} $ & $\textbf{9.8} \pm \textbf{1.2}$ & 48.4
    \\

  \multicolumn{1}{l}{
    confidence head} & $54.4 \pm 0.8$ & $ 3.4 \pm 0.4 $ & 46.4
    \\
    \midrule
  \multicolumn{1}{l}{single outlier detection head}  & $ 99.3 \pm 0.0 $ & $ 15.0 \pm 3.8 $ & N/A 
    \\
    \midrule
  \multicolumn{1}{l}{two heads, no pasting} & $98.9 \pm 0.0$
    & $3.5 \pm 0.6$ & 46.27\\
    \midrule
  \multicolumn{1}{l}{two heads(=LDN\_BIN)}  & $99.3 \pm 0.0$ & $34.9 \pm 6.8$ & \textbf{47.9} 
    \\
  \multicolumn{1}{l}{
    C$\times$ multi-class(=LDN\_OE)} & $ \textbf{99.5} \pm \textbf{0.0}$ & $33.8 \pm 5.1$ & 47.8 
    \\
  \multicolumn{1}{l}{
    C+1$\times$ multi-class}& 
    $98.9 \pm 0.1$ & $25.6 \pm 5.5$ & 46.2
    \\
  \multicolumn{1}{l}{
    C$\times$ multi-label} & $98.8 \pm 0.1$ & $\textbf{49.1}\pm \textbf{5.6}$ & 43.4
    \\
    \bottomrule

\end{tabular}
}
\caption{Validation of dense 
  outlier detection approaches.
  WD denotes WildDash 1 val, MC denotes models
  trained and evaluated using Monte-Carlo dropout.}
\label{table:OOD_detection}
\end{table}

Section 4 of the table 
compares different open-set recognition
modules which train on pasted noisy negatives from
ImageNet-1k-bb as explained 
in \ref{ssec:noisy_negatives}. 
The two-head module architecture
is illustrated in Figure \ref{fig:two-head-module}, while
the other three variants are illustrated 
in Figure \ref{fig:dense_osr_modules}. 
The C-way multi-class approach
trains the model to emit low max-softmax in
outlier samples \cite{liang18iclr,lee18iclr,hendrycks19iclr}
(Figure \ref{fig:oe_module}).
The C+1-way multi-class
model performs prediction over C+1 classes, where
the C+1st class is the outlier class (Figure \ref{fig:c+1-module}).
Finally, the C-way multi-label approach trains C independent
heads with sigmoidal activation 
(Figure \ref{fig:sigmoid_module}).
Comparison with the top section clearly
confirms our hypothesis
that training with diverse noisy negatives
can substantially improve outlier detection. 
We also note a slight reduction 
of the segmentation score
in the column 4.
This reduction is the lowest for 
the C-way multi-class model 
and the two-head model.

A closer inspection of models trained
with noisy negatives shows that the C+1-way
multi-class model performs the worst.
The multi-label model performs 
well on outlier detection
but quite poorly on inlier segmentation.
The two-head model and the C-way multi-class
model perform quite similarly, though further
qualitative analysis shows that they
differ in the type of errors they produce.
The two-head model is more sensitive
to domain shifts between the training and
the validation sets while the C-way multi-class
approach generates false positive
outliers due to low max-softmax score
at semantic borders.

\subsection{Dense open-set recognition on WildDash 1 benchmark}

Table \ref{table:wd_bench_results} presents 
open-set recognition results on
the WildDash 1 benchmark. Our models
are listed in the last three rows of the table.
The LDN\_OE model has 
a single C-way multi-class head
and uses max-softmax for outlier
detection.
The LDN\_BIN and LDN\_BIN$_{\textrm{JS}}$
models have separate heads
for semantic segmentation and
outlier detection. The $JS$ label
indicates training with
scale jittering.
All three models have been trained on
Vistas train, Cityscapes train, 
and WildDash 1 val (inliers) and
ImageNet-1k-bb (noisy negatives).

\setlength{\tabcolsep}{4pt}
\begin{table}[htb]
\begin{center}
\resizebox{\textwidth}{!}{%
\begin{tabular}{l rrrrrrrr}
  \multirow{2}{*}{Model}
  & \multirow{2}{*}{Meta Avg} &
  & \multicolumn{4}{c}{Classic} 
  && \multirow{2}{*}{Negative}
\\
\cmidrule{4-7}
  & mIoU cla
  &
  & mIoU cla
  & iIoU cla
  & mIoU cat
  & iIoU cat
  &
  & mIoU cla\\
\toprule
  \multicolumn{1}{l}{DRN\_MPC \cite{Yu2017}} & 28.3 && 29.1 & 13.9 & 49.2 & 29.2 && 15.9 \\

  \multicolumn{1}{l}{DeepLabv3+\_CS \cite{chen2018encoder}} & 30.6 & & 34.2 & 24.6 & 49.0 & 38.6 && 15.7\\ 

  \multicolumn{1}{l}{MapillaryAI\_ROB \cite{bulo2017place}} &38.9 && 41.3 & 38.0 & 60.5 & 57.6 && 25.0\\

  \multicolumn{1}{l}{AHiSS\_ROB \cite{meletis2018training}} & 39.0 && 41.0 & 32.2 & 53.9 & 39.3 && 43.6\\

  \multicolumn{1}{l}{MSeg \cite{MSeg_2020_CVPR}} & 43.0 && 42.2 & 31.0 & 59.5 & 51.9 && 51.8\\

  \multicolumn{1}{l}{MSeg\_1080 \cite{MSeg_2020_CVPR}} & \textbf{48.3} && \textbf{49.8} & \textbf{43.1} & 63.3 & 56.0 && \textbf{65.0}\\
\midrule
  \multicolumn{1}{l}{LDN\_BIN (ours)} & 41.8 && 43.8 & 37.3 &58.6 & 53.3 && 54.3\\

  \multicolumn{1}{l}{LDN\_OE (ours)} & 42.7 && 43.3 & 31.9 & 60.7 & 50.3 && 52.8\\
\midrule

  \multicolumn{1}{l}{LDN\_BIN$_{\textrm{JS}}$(ours)} & 46.9 && 48.8 & 42.8 & \textbf{63.6} & \textbf{59.3} && 47.7\\
\bottomrule
\end{tabular}
}
\caption{Open-set segmentation
        results on the WildDash 1 benchmark}
\label{table:wd_bench_results}
\end{center}
\end{table}
LDN\_BIN and LDN\_OE differ only
in open-set recognition modules, with the rest of
the training setup being identical.
The two-head model performs 
better in most classic evaluation categories
as well as in the negative category,
however it has a lower meta average score.
This is caused by a larger performance 
drop in most hazard categories 
(more details can be found on 
the WildDash 1 web site).

LDN\_BIN$_{\textrm{JS}}$ has the same
architecture as LDN\_BIN but it is trained
using scale jittering
to be able to perform inference
on larger resolutions
(768x1451). 
This setup
improves the segmentation accuracy
across all categories
and reduces sensitivity to hazards
while slightly deteriorating performance
in negative images. 
We did not retrain
LDN\_OE using scale jittering since 
this model produces
false positives on semantic borders
regardless of the inference resolution.

The best overall performance is achieved
by the MSeg\_1080 \cite{MSeg_2020_CVPR}.
However, that model uses much more
negative supervision:
densely labeled Ade20k and COCO (they) vs 
bounding boxes from ImageNet-1k (us). 
Additionally, they train and evaluate
on a larger resolution (1024 vs 768) and use
a model with almost 4 times more parameters
(65.8M vs 17.4M ).
Mseg\_1080 is somewhat less sensitive to some hazards
(most significantly underexposure) which
may be due to a significantly
larger inlier training dataset. 
Aside from Vistas and Cityscapes, they also use
BDD (8000 images) and
IDD (7974 images).
On the other hand, MSeg does not use 70
images from WildDash 1 val.
Our model is competitive
and actually outperforms MSeg when evaluated on 
the same resolution (MSeg vs LDN\_BIN).

\begin{figure}[htb]
  \centering
  \includegraphics[width=\columnwidth]{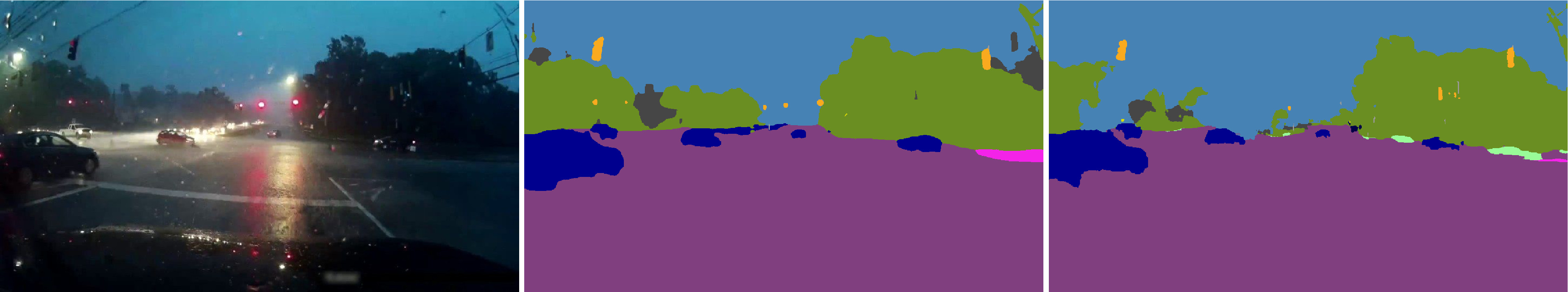}
  \\[0.2em]
  \includegraphics[width=\columnwidth]{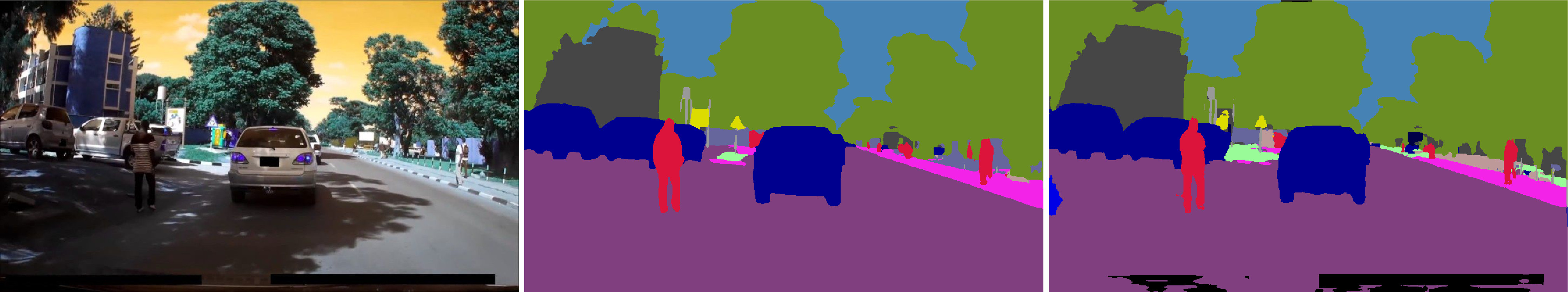}
  \\[0.2em]
  \includegraphics[width=\columnwidth]{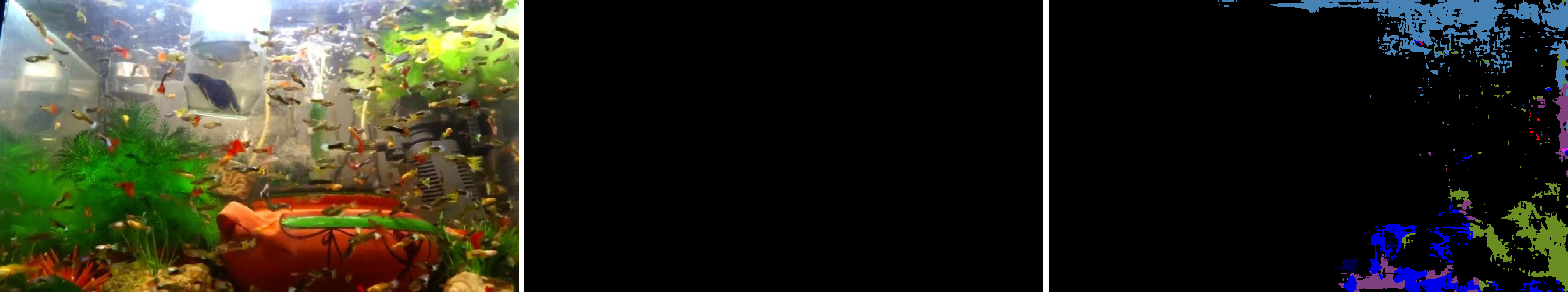}
  \caption{Qualitative comparison between
  MSeg (middle column) and LDN\_BIN$_{\textrm{JS}}$ (right column)
  on WildDash 1 test images (left column).
  MSeg performs better on some negative
  images (row 3), and small objects (row 1),
  but it appears unable to locate outlier patches 
  in front of inlier background (row 2).
  }
  \label{fig:bench_mseg_ldn}
\end{figure}

Figure \ref{fig:bench_mseg_ldn} presents
a qualitative comparison between MSeg and
LDN\_BIN$_{\textrm{JS}}$ as shown 
on the WildDash 1 benchmark.
The columns show: i) original image, 
ii) MSeg output and 
iii) LDN\_BIN$_{\textrm{JS}}$ output.
Images show that MSeg
performs better on small objects
and negative images
which is likely due to larger resolution
and more supervision.
Note however that the MSeg model does not recognize
black rectangles (row 2) as outliers.
Detailed qualitative results for
LDN\_BIN and LDN\_OE
can be found in \cite{bevandic19gcpr}.

\subsection{Open-set validation on Lost and Found dataset}
\label{ss:fishyval}

Table \ref{table:val_fishy_outlier} shows 
evaluation on the validation subset of Fishyscapes 
Lost and Found.
All models were trained on inliers from
Vistas train, Cityscapes train, and WildDash 1 val. 
LDN$_{\textrm{JS}}$ denotes the max-softmax baseline 
trained with scale jittering and without outliers.
All other models were also trained 
on noisy negatives from ImageNet-1k-bb.
LDN\_OE, LDN\_BIN and 
LDN\_BIN$_{\textrm{JS}}$
are exact same models 
we submitted to the WildDash 1 benchmark.
LDN\_BIN$_{\textrm{JS, RSP}}$ 
has the same architecture 
as LDN\_BIN$_{\textrm{JS}}$,
however it varies the size of
pasted negatives during
training in order to improve
detection of smaller outliers.
The last row combines
our OOD head with max-softmax using
multiplication.
Later we show that this formulation succeeds
since max-softmax complements our method
when the outliers are very small

\setlength{\tabcolsep}{4pt}
\begin{table}[htb]
\begin{center}
\begin{tabular}{llrrrr}
  Model & criterion &AP & AUROC & FPR95 & mIoU\\
\toprule
    LDN$_{\textrm{JS}}$ & MSM & 7.8 & 92.1 & 26.6 & 76.4\\
\midrule
   LDN\_OE & MSM & 9.5 & 88.8 & 44.2 & 72.2 \\

    LDN\_BIN & OP & 13.2 & 88.0 & 71.9 & 75.1\\

  LDN\_BIN$_\textrm{{JS}}$ & OP & 25.4 & 89.8 & 90.0 & \textbf{76.5}\\
[0.5em]
  LDN\_BIN$_{\textrm{JS, RSP}}$ & OP & 36.9 & \textbf{96.1} & \textbf{20.0} & 76.3\\

  & OP $\times$ MSM & \textbf{45.7} & 95.6 & 24.0 &\\
\bottomrule
\end{tabular}
\caption{Comparison of open-set segmentation approaches 
  on Fishyscapes Lost and Found (AP, AUROC, FPR95(\%))
  and Vistas (mIoU) validation subsets. MSM is short for max-softmax,
  while OP stands for outlier probability estimated
  by the outlier detection head.
}
\label{table:val_fishy_outlier}
\end{center}
\end{table}
\setlength{\tabcolsep}{1.4pt}

Both anomaly detection
and semantic segmentation benefit from 
scale jittering during training.
This is different than WildDash 1
where scale jittering decreased performance
on negative images (cf. Table \ref{table:wd_bench_results}).

\begin{figure}[htb]
\centering
\begin{tabular}{cccc}
\multicolumn{2}{c}{LDN$_{\textrm{JS}}$}&\multicolumn{2}{c}{LDN\_BIN$_{\textrm{JS, RSP}}$}\\

\includegraphics[width=0.25\columnwidth]{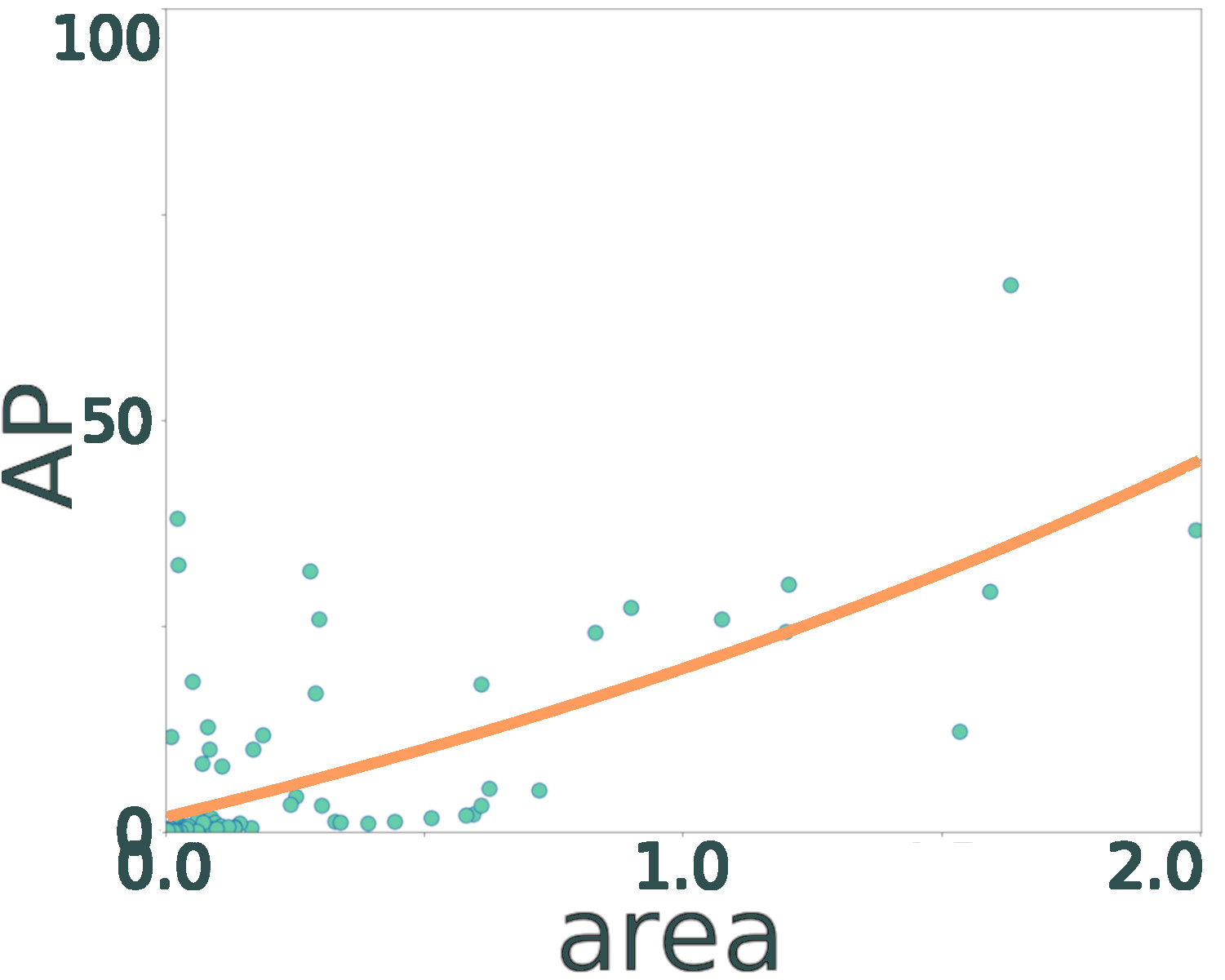}
&\includegraphics[width=0.25\columnwidth]{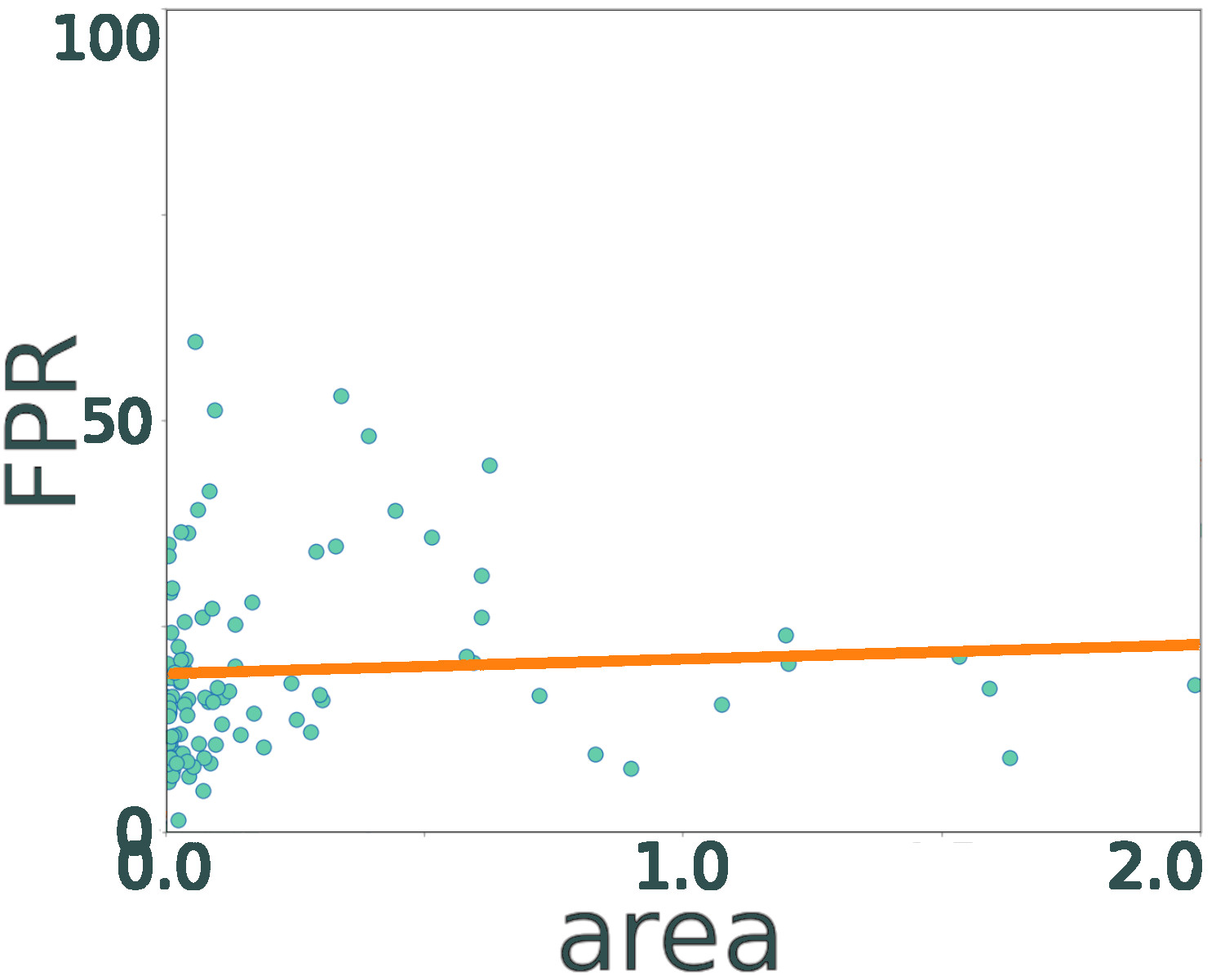}
&\includegraphics[width=0.25\columnwidth]{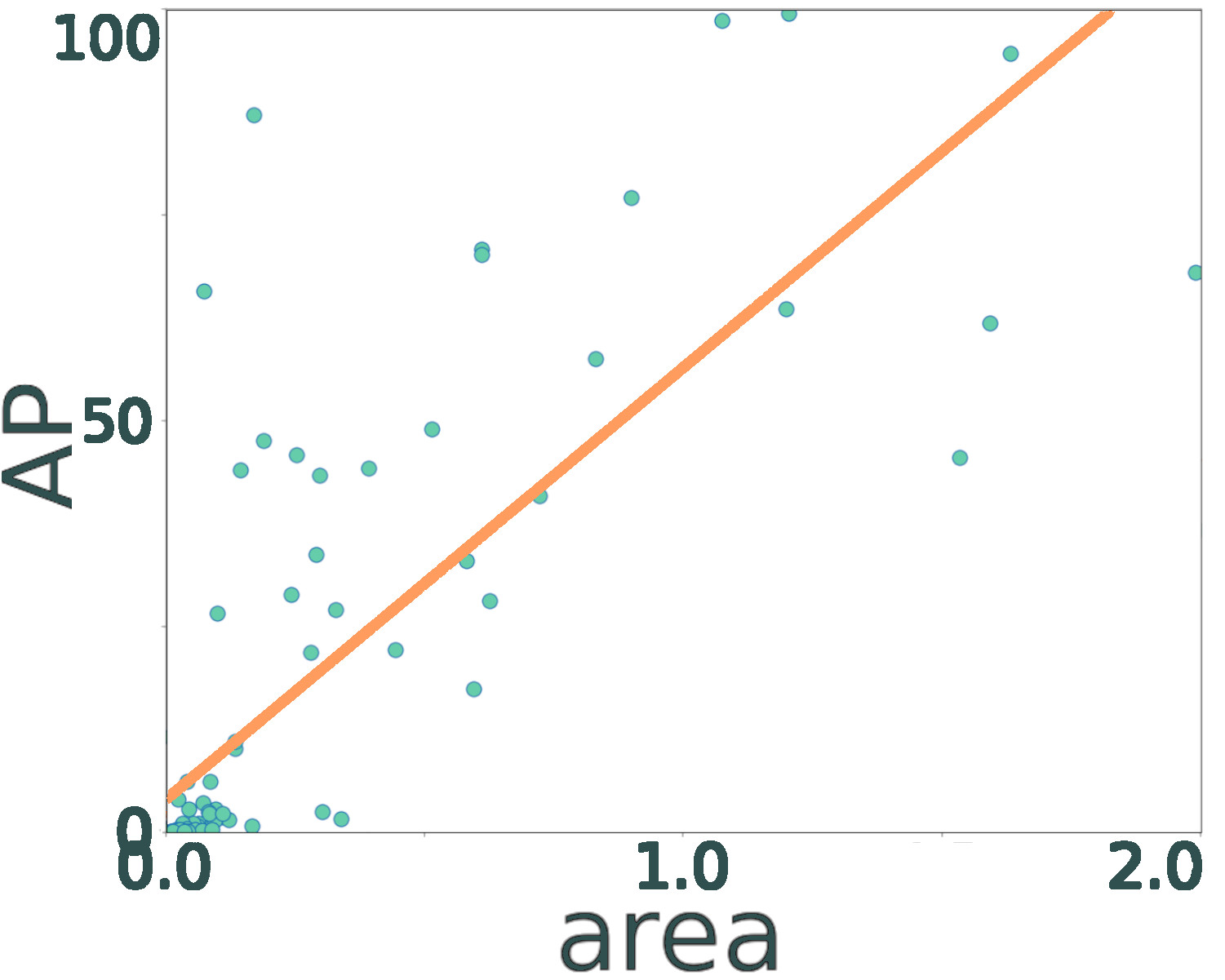}
&\includegraphics[width=0.25\columnwidth]{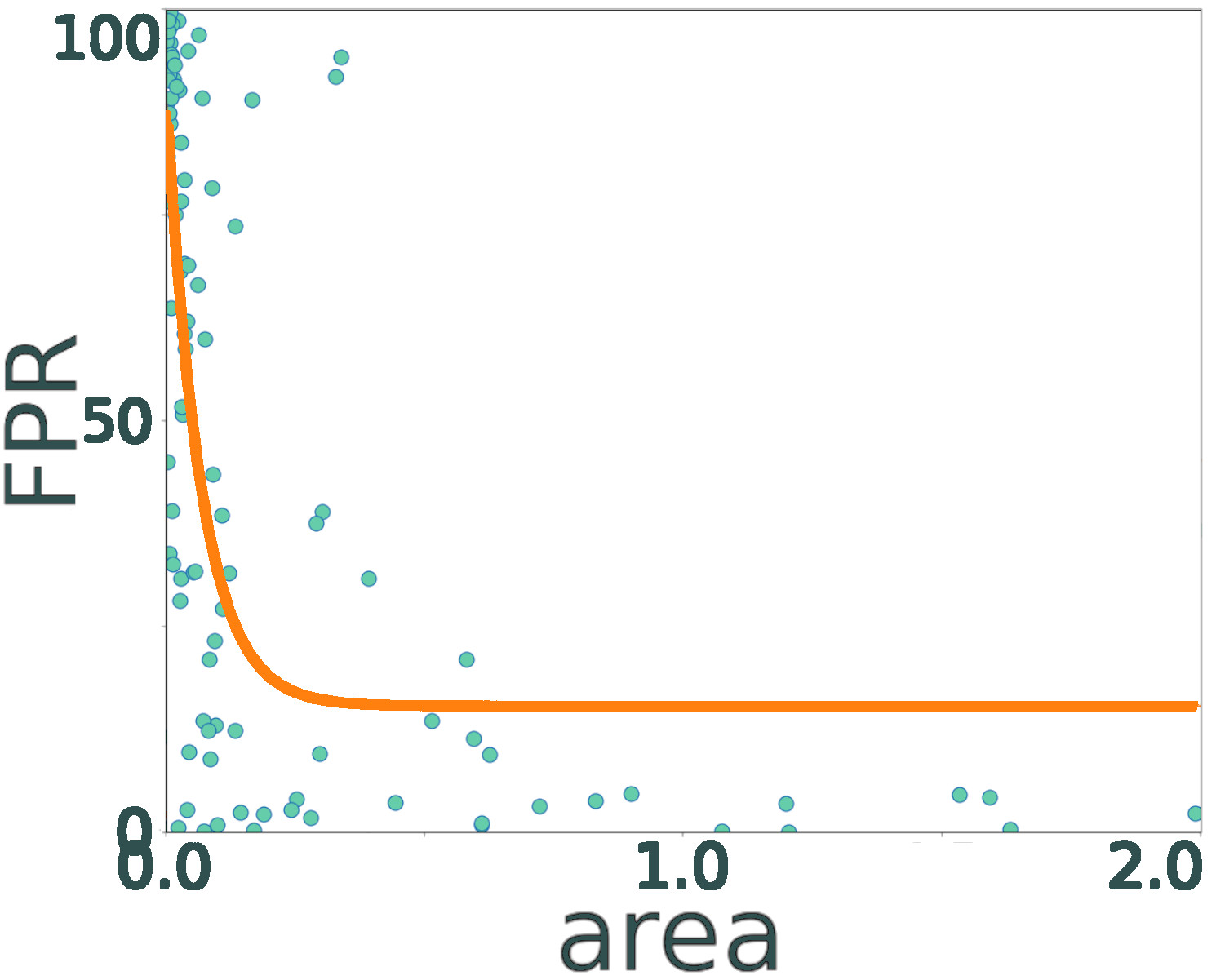}
\end{tabular}\\
\caption{Influence of the outlier size on the model performance
on Fishyscapes Lost and Found val. 
The two leftmost graphs show AP and FPR95
of the max-softmax baseline (LDN$_{\textrm{JS}}$)
and the two rightmost graphs show AP and FPR95
for our model trained with noisy negatives
(LDN\_BIN$_{\textrm{JS, RSP}}$). 
Higher AP and lower FPR scores indicate 
that our model prevails on 
large outliers.
Max-softmax on the other hand achieves better
results on small outliers because it detects
object edges well.}
\label{fig:fishy_val_per_image}
\end{figure}

Figure \ref{fig:fishy_val_per_image} explores the influence
of outlier size on model performance
by plotting the relation
between the outlier area and
the detection performance.
The figure shows AP and FPR95
with respect to the area of the outlier object
for LDN$_{\textrm{JS}}$
(which uses max-softmax for outlier detection),
and LDN\_BIN$_{\textrm{JS, RSP}}$.
We see that the accuracy of both models
depend on the size of the outlier.
Max-softmax acts as an edge detector and
therefore performs better 
on smaller objects. It however performs
poorly on larger objects because it is unable
to detect the interior of an object as an outlier.

Figure \ref{fig:fishy_val_per_image} implies that we
can improve the accuracy of our
two head models on small objects by 
multiplying the  outlier probability
with max-softmax.
\begin{myalign}
  \label{eq:comb}
  P(outlier_{ij}|x) =
    P(o_{ij}=1|x)\cdot(1-max_{c}(P(y_{ijc}|x))
\end{myalign}
This equation suggests that outliers should both
appear strange to outlier detection
head \emph{and} produce small max-softmax scores.
This formulation improves upon max-softmax by dampening
outlier probabilities on semantic borders, since
our trained outlier detection head perceives them
as iniliers. This formulation improves upon our 
trained outlier detection head on small outliers, since
that is where max-softmax achieves fair performance.

Note that relatively poor performance of our
model on small outliers does not come as 
a great surprise. Our predictions are 4 
times subsampled with respect to the input
resolution to reduce computational
complexity and memory footprint during
training. This is a common
trade-off \cite{russakovsky15ijcv}
which can be avoided, but
at a great computational cost
\cite{DBLP:zhu18cvpr}.

\begin{figure}[htb]
  \centering
  \includegraphics[width=\columnwidth]{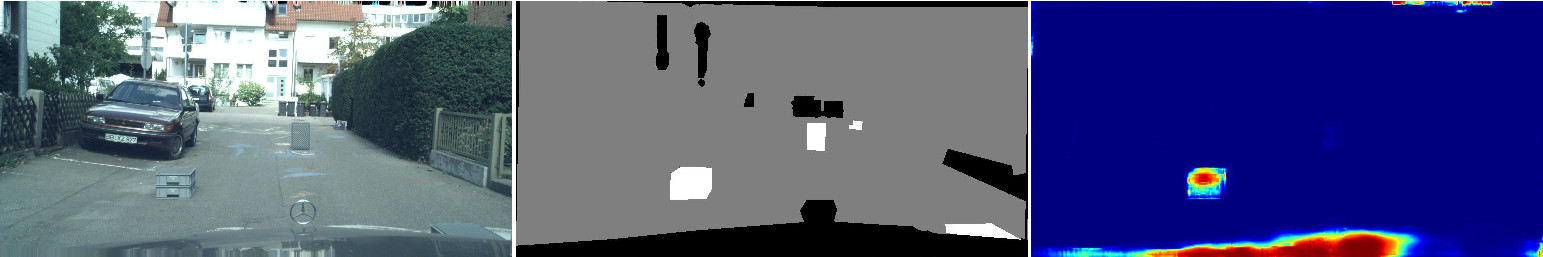}
  \\[0.2em]
  \includegraphics[width=\columnwidth]{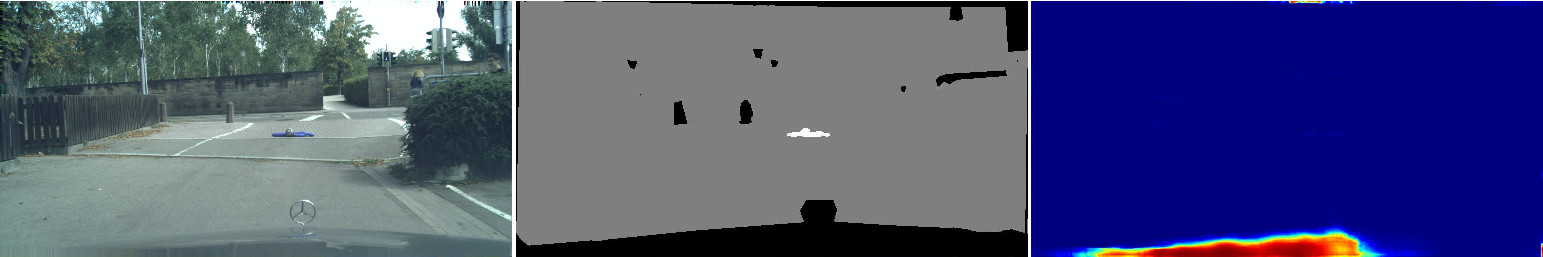}
  \\[0.2em]
  \includegraphics[width=\columnwidth]{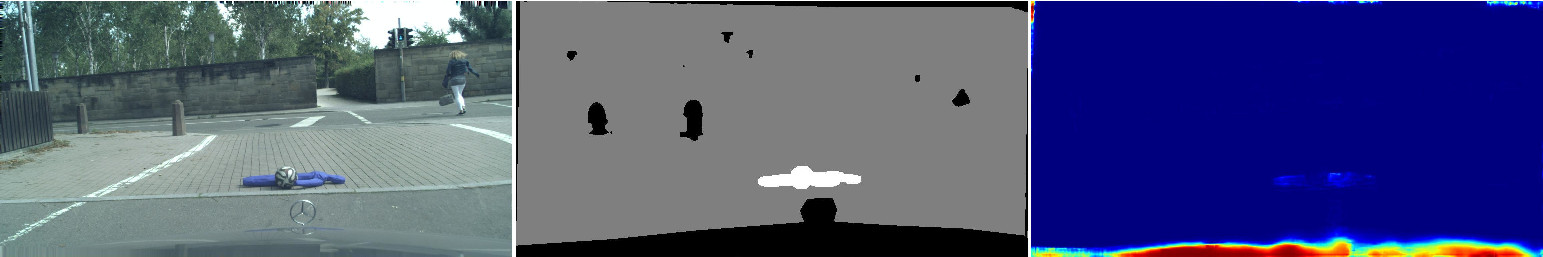}

  \caption{Outlier detection with LDN\_BIN$_{JS, RSP}$
  and OP$\times$MSM
  on Fishyscapes Lost and Found val.
  Columns present i) the original image,
  ii) the ground truth labels, and 
  iii) the outlier probability. 
  Our model works better on close objects 
  than on distant ones (row 1). 
  The outlier detection confidence grows 
  as the camera draws nearer (rows 2, 3).
  Very small outliers are not detected (rows 1, 2).
  }
  \label{fig:fishy_val_results}
\end{figure}

Figure \ref{fig:fishy_val_results} shows 
qualitative performance of our model.
Column 1 presents the original image.
Column 2 contains the ground truth, with
inlier, outlier and ignore pixels denoted in 
gray, white and black respectively. 
Finally, column 3 shows the output of our
LDN\_BIN$_{JS,RSP}$ model using
a conjunction between our prediction and 
max-softmax probability (OD$\times$MSP).
Our model performs
well on larger and closer objects (rows 1 and 3),
while struggling with distant and small objects (rows 1 and 2).
Finally, we note that some of
the ignore pixels (e.g. ego-vehicle, noise
on image borders) are also 
classified as anomalies. 

\subsection{Dense open-set recognition on the Fishyscapes benchmark}
Table \ref{table:fishy_bench_results} shows current
results on the Fishyscapes benchmark \cite{blum19iccvw}.
The benchmark provides segmentation accuracy 
on Cityscapes val, as well as
outlier detection accuracy on 
FS Lost and Found and FS Static.
FS Lost and Found comprises 300 images taken from 
the Lost and Found dataset. These images are
relabelled to distinguish
between inlier, outlier and void classes, and 
filtered to exclude road hazards which correspond to
inlier classes (e.g. bicycles).
FS static was created by pasting
PASCAL VOC objects into Cityscapes images.

Note that LDN\_BIN$_{\textrm{JS}}$ 
is almost exactly the same model that was 
presented in Table \ref{table:wd_bench_results}. 
Due to the requirement of the benchmark,
the model had to be reimplemented in Tensorflow 1.
We did not retrain the model, but reused the
parameters learnt in Pytorch.

As in our validation experiments (see \ref{ss:fishyval}), 
LDN\_BIN$_{\textrm{JS, RSP}}$
improves the detection of smaller outliers. 
This is reflected by an improved FPR95 score 
with respect to LDN\_BIN$_{\textrm{JS}}$.
The best results was achieved with 
LDN\_BIN$_{\textrm{JS, RSP, ADE20K}}$
where instances from the ADE20K dataset
were used as negatives during training.
We outperform other models by a large
margin on FS static.
We also achieve the best FPR95 and a close
second-best outlier detection AP on Lost and Found
without significant drop in 
segmentation performance that 
occurs in the best submission.

\setlength{\tabcolsep}{4pt}
\begin{table*}[htb]
\begin{center}
\resizebox{\textwidth}{!}{%
\begin{tabular}{lllccrrrrrrr}

  \multicolumn{2}{l}{\multirow{2}{*}{Model}} & \multirow{2}{*}{Criterion} & \multirow{2}{*}{Train} & \multirow{2}{*}{OoD} & \multirow{2}{*}{City} && \multicolumn{2}{c}{Lost and Found} && \multicolumn{2}{c}{FS Static}\\
\cmidrule{8-9}\cmidrule{11-12}
  &  & & & & mIoU && AP &  FPR95 && AP & FPR95 \\
\toprule
   \multicolumn{2}{l}{Dirichlet DeepLab \cite{blum19iccvw}} & prior entropy & \ding{51} & \ding{51} & 70.5 && \textbf{34.3} & 47.4 && 31.3 & 84.6 \\

    \multicolumn{2}{l}{Bayesian DeepLab \cite{blum19iccvw}}& mutual information & \ding{51} & \ding{55} & 73.8 && 9.8 &  38.5 && 48.7 & 15.5 \\
         OoD training \cite{blum19iccvw} & &maximize entropy & \ding{51} & \ding{51} & 79.0 && 1.74 & 30.6 && 27.5 & 23.6 \\
     [0.5em]
  \multicolumn{2}{l}{Softmax \cite{blum19iccvw}} & entropy & \ding{55} & \ding{55} & 80.0 && 2.9 &  44.8 && 15.4 & 39.8 \\

     & &max-softmax (MSM) & \ding{55} & \ding{55} &  && 1.8 &  44.9 && 12.9 & 39.8 \\
[0.5em]
     \multicolumn{2}{l}{Learned embedding density \cite{blum19iccvw}} & logistic regression & \ding{55} & \ding{51} & 80.0 && 4.7 &  24.4 && 57.2 & 13.4 \\
 
      & &  minimum nll & \ding{55} & \ding{55} & && 4.3 &  47.2 && 62.1 & 17.4 \\

     & &single-layer nll & \ding{55} & \ding{55} & && 3.0 &  32.9 && 40.9 & 21.3 \\
[0.5em]

     \multicolumn{2}{l}{Image resynthesis} & resynthesis difference & \ding{55} & \ding{55} & \textbf{81.4} && 5.7 &  48.1 && 29.6 & 27.1 \\
\midrule
  \multirow{3}{3cm}{Discriminative outlier detection head (ours)}& \multicolumn{1}{l}{LDN\_BIN$_{\textrm{JS}}$} & outlier probability (OP) & \ding{51} & \ding{51} & 77.7 && 15.7 &  76.9 && 82.9 & 5.1 \\
      [0.5em]
    & \multicolumn{1}{l}{LDN\_BIN$_{\textrm{JS, RSP}}$} & outlier probability (OP) & \ding{51} & \ding{51} & 77.3 && 21.2 &  36.9 && 86.2 & 2.4 \\
   & \multicolumn{1}{l}{} & OP $\times$ MSM & \ding{51} & \ding{51} && & 30.9 &  22.2 && 84.0 & 10.3 \\
	& \multicolumn{1}{l}{LDN\_BIN$_{\textrm{JS, RSP, ADE20K}}$} & OP $\times$ MSM & \ding{51} & \ding{51} && & 31.3 &  \textbf{19.0} && \textbf{96.8} & \textbf{0.3} \\
\bottomrule
\end{tabular}
}
\caption{Open-set segmentation evaluation 
 on the Fishyscapes benchmark.}
 \label{table:fishy_bench_results}
\end{center}
\end{table*}
\setlength{\tabcolsep}{1.4pt}

\subsection{Open-set segmentation on StreetHazard}
Table \ref{table:caos_results} presents
open-set segmentation accuracy 
on StreetHazard.
We evaluate the same models as in previous experiments
(LDN$_{\textrm{JS}}$, LDN\_BIN$_{\textrm{JS}}$
 and LDN\_BIN$_{\textrm{JS, RSP}}$)
and compare them with the max-softmax baseline.
We ignore outlier pixels when measuring
segmentation accuracy. 
Unlike
\cite{hendrycks2019anomalyseg},
we do not use ignore pixels during evaluation 
(same as \cite{blum19iccvw}).
Furthermore, we do not report 
the mean of per-image scores.
In our view, such practice may yield
over-optimistic estimate of the overall
anomaly detection metrics, since recognition
errors can not propagate across images.
We therefore determine global scores
on 10 times downsampled predictions.
We evaluated the performance by measuring
the mean of per-image scores and obtained 
similar results to the ones we report.

\begin{figure}[htb]
  \centering
  \includegraphics[width=0.95\columnwidth]{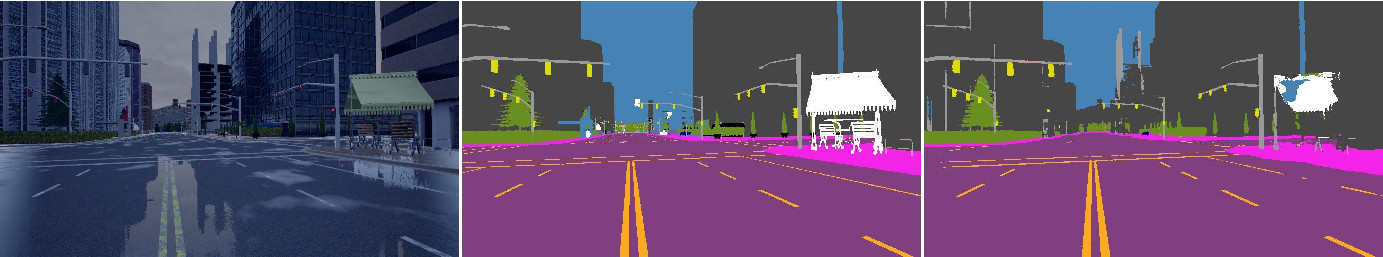}
  \\[0.2em]
  \includegraphics[width=0.95\columnwidth]{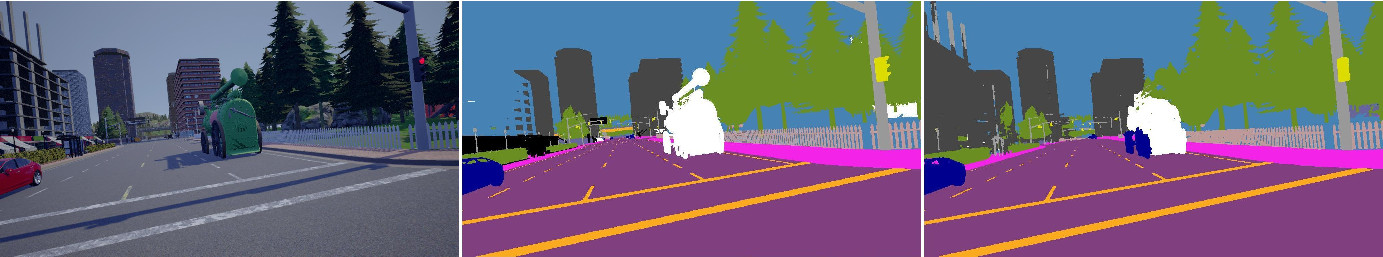}
  \\[0.2em]
  \includegraphics[width=0.95\columnwidth]{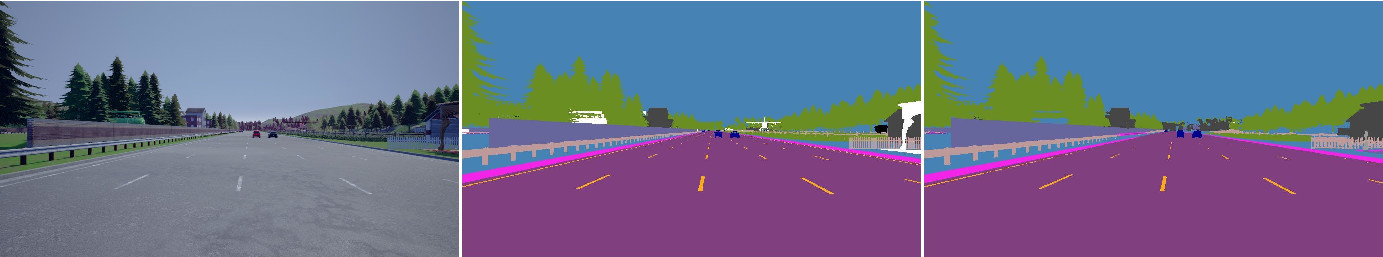}
  \caption{Open-set segmentation  
  on StreetHazard.
  The columns show the input
  image, ground truth segmentation
  and the output of LDN\_BIN$_{\textrm{JS, RSP}}$. 
  Outliers are white while ignore pixels are black.
  Our model performs better on
  large outliers (rows 1, 2)
  than on small ones (row 3).
  }
  \label{fig:sh_np_results}
\end{figure}

\setlength{\tabcolsep}{4pt}
\begin{table}[htb]
\begin{center}
\resizebox{\textwidth}{!}{%
\begin{tabular}{llrrrr}

   model & criterion & AP & AUROC & FPR95 & test mIoU
 \\
\toprule
  PSPNet \cite{hendrycks2019anomalyseg}  & CRF+msm & 6.5 & 88.1 & 29.9 & N/A \\

   PSPNet \cite{franchi2019tradi} & TRADI & 7.2 & 89.2 & \textbf{25.3} & N/A\\

\midrule
   LDN$_{\textrm{JS}}$ & MSM & 7.28 & 87.63 & 38.13 & 65.04\\

   LDN\_BIN$_{\textrm{JS}}$ & OP & 18.56 & 87.00 & 79.08 & 66.32\\
[0.5em]
   LDN\_BIN$_{\textrm{JS, RSP}}$ & OP & \textbf{19.74} & 88.86 & 56.19 & \textbf{66.94} \\
    & OP$\times$MSM & 18.82 & \textbf{89.72} & 30.86 & \\
 \bottomrule
\end{tabular}
}
\caption{Performance evaluation
on StreetHazard}
\label{table:caos_results}
\end{center}
\end{table}
\setlength{\tabcolsep}{1.4pt}
Figure \ref{fig:sh_np_results}
shows some qualitative results.
The columns represent: i) the original image, 
ii) the ground truth and iii) our output.

\section{Conclusion}

We have presented a novel approach
for dense outlier detection
and open-set recognition.
The main idea is to discriminate
an application-specific inlier dataset
(e.g.\ Vistas, Cityscapes),
from a diverse general-purpose dataset
(e.g.\ ImageNet-1k).
Pixels from the latter dataset
represent noisy test-agnostic negative samples.
We train on mixed batches
with approximately equal share
of inliers and noisy negatives.
This promotes robustness
to occasional inliers in negative images
and favours stable development
of batchnorm statistics.
We encourage correct recognition
of spatial borders
between outlier and inlier pixels
by pasting negative patches
at random locations in inlier images.
Consequently, the resulting models
succeed to generalize
in test images with mixed content.
We have shown that feature sharing
greatly improves dense outlier detection,
while only slightly deteriorating
semantic segmentation.
The resulting multi-task architecture
is able to perform dense open-set recognition
with a single forward pass.

This is the first and currently the only method
which competes at both dense
open-set recognition benchmarks, Fishyscapes
and WildDash 1.
Currently, our model is at the top
on Fishyscapes Static leaderboard,
and a close runner-up on WildDash 1
while training with less supervision
than the top rank algorithm
\cite{MSeg_2020_CVPR}.
The same model also achieves
the runner-up AP and competitive FPR 95
on Fishyscapes Lost and Found.
We achieve state-of-the-art AP accuracy on the
StreetHazard dataset despite a 
strong domain shift between our negative
dataset (ImageNet-1k-bb) and the test
dataset.

Our method outperformed
the max-softmax baseline in all experiments.
The advantage is greatest when the outliers are large,
such as in Fishyscapes static and WildDash 1.
A conjunction of our method and max-softmax
becomes advantageous on Fishyscapes Lost and Found.
This suggests that our method and max-softmax
target independent traits of outlier pixels.
Most reported experiments
feature the same model,
hyper parameters,
training procedure,
and the negative dataset:
only the inliers are different.

The reported results
confirm our hypotheses
i) that noisy negatives can improve
dense outlier detection
and open-set recognition,
and ii) that the shared features
greatly improve outlier detection 
without significant deterioration
of semantic segmentation. The resulting
open-set models
perform comparably with respect
to their closed-set counterparts.
Suitable directions
for future work include
improving our models on small
outliers,
as well as incorporating joint training
with generative models.

\section*{Acknowledgment}
This  work  has  been  supported  
by the Croatian Science Foundation under
the grant ADEPT and the  European Regional
Development  Fund  
under  the  grant    
DATACROSS.

\section*{Published Paper}
This manuscript has been accepted for publication
Image and Vision Computing,
after peer review and is subject to their terms of use,
but is not the Version of Record and does not 
reflect post-acceptance improvements, or any corrections.
The Version of Record is available here:
https://www.sciencedirect.com/science/article/pii/S0262885622001196.







\bibliographystyle{elsarticle-num-names}
\bibliography{sample.bib}







\end{document}